\documentclass{article}
\usepackage[letterpaper, left=1in, right=1in, top=1in, bottom=1in]{geometry}
\usepackage{natbib}
\bibpunct{(}{)}{;}{a}{,}{,}
\usepackage[colorlinks,citecolor=blue!70!black,linkcolor=blue!70!black,
urlcolor=blue!70!black,breaklinks=true]{hyperref}
\usepackage{parskip}
\usepackage[usenames, dvipsnames]{xcolor}
\usepackage{microtype}
\usepackage{booktabs} 
\usepackage{amsmath}
\usepackage{dsfont} 
\usepackage{amssymb}
\usepackage{xspace}
\usepackage{url}            
\usepackage{algorithm}
\usepackage[noend]{algorithmic}
\usepackage{amsthm}
\usepackage{listings}
\usepackage{bm}
\usepackage{bbm}
\usepackage{enumitem}
\usepackage{nicefrac}       
\usepackage{mathrsfs} 
\usepackage{inconsolata}
\usepackage{wrapfig}
\usepackage{graphicx}
\usepackage{bigints}
\usepackage{transparent}
\usepackage{comment}
\usepackage[nameinlink,capitalize]{cleveref}
\makeatletter
\AddToHook{cmd/appendix/before}{
\def\cref@section@alias{appendix}
\def\cref@subsection@alias{appendix}
\def\cref@subsubsection@alias{appendix}
}
\makeatother
\usepackage{thmtools}
\usepackage{thm-restate}
\usepackage{nicematrix}
\usepackage{arydshln}
\usepackage{fancyhdr}
\usepackage{mathtools}
\usepackage[scaled=.88]{helvet}
\usepackage{breakcites}
\usepackage{multirow}
\usepackage{booktabs}
\usepackage{threeparttable}
\usepackage[normalem]{ulem}
\usepackage{graphicx}
\usepackage{subcaption}

\renewcommand{\epsilon}{\varepsilon}

\newtheoremstyle{spaced}
  {6pt}
  {0pt}
  {\itshape}
  {}
  {\bfseries}
  {.}
  {0.5em}
  {}
\theoremstyle{spaced}

\newcommand{\algcommentlight}[1]{\textcolor{blue!70!black}{\transparent{0.5}\small{\texttt{\textbf{//\hspace{2pt}#1}}}}}

\DeclarePairedDelimiterX{\infdiv}[2]{(}{)}{%
  #1\;\delimsize\|\;#2%
}

\def\ddefloop#1{\ifx\ddefloop#1\else\ddef{#1}\expandafter\ddefloop\fi}
\def\ddef#1{\expandafter\def\csname bb#1\endcsname{\ensuremath{\mathbb{#1}}}}
\ddefloop ABCDEFGHIJKLMNOPQRSTUVWXYZ\ddefloop
\def\ddefloop#1{\ifx\ddefloop#1\else\ddef{#1}\expandafter\ddefloop\fi}
\def\ddef#1{\expandafter\def\csname b#1\endcsname{\ensuremath{\mathbf{#1}}}}
\ddefloop ABCDEFGHIJKLMNOPQRSTUVWXYZ\ddefloop
\def\ddef#1{\expandafter\def\csname sf#1\endcsname{\ensuremath{\mathsf{#1}}}}
\ddefloop ABCDEFGHIJKLMNOPQRSTUVWXYZ\ddefloop
\def\ddef#1{\expandafter\def\csname c#1\endcsname{\ensuremath{\mathcal{#1}}}}
\ddefloop ABCDEFGHIJKLMNOPQRSTUVWXYZ\ddefloop
\def\ddef#1{\expandafter\def\csname h#1\endcsname{\ensuremath{\widehat{#1}}}}
\ddefloop ABCDEFGHIJKLMNOPQRSTUVWXYZ\ddefloop
\def\ddef#1{\expandafter\def\csname hc#1\endcsname{\ensuremath{\widehat{\mathcal{#1}}}}}
\ddefloop ABCDEFGHIJKLMNOPQRSTUVWXYZ\ddefloop
\def\ddef#1{\expandafter\def\csname t#1\endcsname{\ensuremath{\widetilde{#1}}}}
\ddefloop ABCDEFGHIJKLMNOPQRSTUVWXYZ\ddefloop
\def\ddef#1{\expandafter\def\csname tc#1\endcsname{\ensuremath{\widetilde{\mathcal{#1}}}}}
\ddefloop ABCDEFGHIJKLMNOPQRSTUVWXYZ\ddefloop
\def\ddefloop#1{\ifx\ddefloop#1\else\ddef{#1}\expandafter\ddefloop\fi}
\def\ddef#1{\expandafter\def\csname scr#1\endcsname{\ensuremath{\mathscr{#1}}}}
\ddefloop ABCDEFGHIJKLMNOPQRSTUVWXYZ\ddefloop

\let\oldparagraph\paragraph
\renewcommand{\paragraph}[1]{\oldparagraph{#1}}

\renewcommand{\epsilon}{\varepsilon}

\renewcommand{\bigm}[1]{%
  \ifcsname fenced@\string#1\endcsname
    \expandafter\@firstoftwo
  \else
    \expandafter\@secondoftwo
  \fi
  {\expandafter\amsmath@bigm\csname fenced@\string#1\endcsname}%
  {\amsmath@bigm#1}%
}

\newcommand{\DeclareFence}[2]{\@namedef{fenced@\string#1}{#2}}
\makeatother

\makeatletter
\let\save@mathaccent\mathaccent
\newcommand*\if@single[3]{%
  \setbox0\hbox{${\mathaccent"0362{#1}}^H$}%
  \setbox2\hbox{${\mathaccent"0362{\kern0pt#1}}^H$}%
  \ifdim\ht0=\ht2 #3\else #2\fi
  }
\newcommand*\rel@kern[1]{\kern#1\dimexpr\macc@kerna}
\newcommand*\widebar[1]{\@ifnextchar^{{\wide@bar{#1}{0}}}{\wide@bar{#1}{1}}}
\newcommand*\wide@bar[2]{\if@single{#1}{\wide@bar@{#1}{#2}{1}}{\wide@bar@{#1}{#2}{2}}}
\newcommand*\wide@bar@[3]{%
  \begingroup
  \def\mathaccent##1##2{%
    \let\mathaccent\save@mathaccent
    \if#32 \let\macc@nucleus\first@char \fi
    \setbox\z@\hbox{$\macc@style{\macc@nucleus}_{}$}%
    \setbox\tw@\hbox{$\macc@style{\macc@nucleus}{}_{}$}%
    \dimen@\wd\tw@
    \advance\dimen@-\wd\z@
    \divide\dimen@ 3
    \@tempdima\wd\tw@
    \advance\@tempdima-\scriptspace
    \divide\@tempdima 10
    \advance\dimen@-\@tempdima
    \ifdim\dimen@>\z@ \dimen@0pt\fi
    \rel@kern{0.6}\kern-\dimen@
    \if#31
      \overline{\rel@kern{-0.6}\kern\dimen@\macc@nucleus\rel@kern{0.4}\kern\dimen@}%
      \advance\dimen@0.4\dimexpr\macc@kerna
      \let\final@kern#2%
      \ifdim\dimen@<\z@ \let\final@kern1\fi
      \if\final@kern1 \kern-\dimen@\fi
    \else
      \overline{\rel@kern{-0.6}\kern\dimen@#1}%
    \fi
  }%
  \macc@depth\@ne
  \let\math@bgroup\@empty \let\math@egroup\macc@set@skewchar
  \mathsurround\z@ \frozen@everymath{\mathgroup\macc@group\relax}%
  \macc@set@skewchar\relax
  \let\mathaccentV\macc@nested@a
  \if#31
    \macc@nested@a\relax111{#1}%
  \else
    \def\gobble@till@marker##1\endmarker{}%
    \futurelet\first@char\gobble@till@marker#1\endmarker
    \ifcat\noexpand\first@char A\else
      \def\first@char{}%
    \fi
    \macc@nested@a\relax111{\first@char}%
  \fi
  \endgroup
}
\makeatother

\usepackage{tcolorbox}
\usepackage{pgfplots}
\usepgfplotslibrary{statistics}

\pgfplotsset{compat=1.18}
\usepackage{tikz}
\usetikzlibrary{pgfplots.groupplots}

\newcommand{\coveragemath}{\mathsf{Coverage}}

\title{Adaptive Test-Time Compute Allocation with \\ Evolving In-Context Demonstrations}
\date{}

\author{
Bowen Zuo$^{1}$\\
{\small\texttt{bzuo002@ucr.edu}}
\and
Dongruo Zhou$^{2}$\\
{\small\texttt{dz13@iu.edu}}
\and
Yinglun Zhu$^{1}$\\
{\small\texttt{yzhu@ucr.edu}}
\and
~\\
{\normalsize $^{1}$University of California, Riverside \quad \quad $^{2}$Indiana University Bloomington}
}

\begin{document}

\maketitle

\begin{abstract}
  While scaling test-time compute can substantially improve model performance, existing approaches either rely on static compute allocation or sample from fixed generation distributions. In this work, we introduce a test-time compute allocation framework that jointly adapts where computation is spent and how generation is performed. Our method begins with a warm-up phase that identifies easy queries and assembles an initial pool of question-response pairs from the test set itself. An adaptive phase then concentrates further computation on unresolved queries while reshaping their generation distributions through evolving in-context demonstrations---conditioning each generation on successful responses from semantically related queries rather than resampling from a fixed distribution. Experiments across math, coding, and reasoning benchmarks demonstrate that our approach consistently outperforms existing baselines while consuming substantially less inference-time compute.

\end{abstract}

\section{Introduction}
\label{sec:intro}

Test-time strategies \citep{wei2022chain, yao2023tree, madaan2023selfrefine} have long been a central focus in the study of Large Language Models (LLMs). 
In recent years, test-time scaling 
\citep{agarwal2024manyshot, muennighoff2025s1, liu2025can, snell2025scaling, brown2024largelanguagemonkeysscaling}
has emerged
as an effective alternative to traditional training-time scaling 
\citep{10.5555/3648699.3648939, 10.5555/3600270.3602446, kaplan2020scalinglawsneurallanguage}.
Rather than increasing model size or training data, test-time scaling improves performance by allocating additional computation at inference time, typically by generating multiple candidate responses and selecting the most promising one.
Representative approaches include Best-of-$N$ sampling \citep{brown2024largelanguagemonkeysscaling, snell2025scaling}, self-consistency \citep{wang2023selfconsistency}, and reward-guided selection using learned or ground-truth verifiers
\citep{zhang-etal-2025-lessons, lightman2024lets,cobbe2021trainingverifierssolvemath, uesato2022solving}.
More recently, reasoning-oriented models
\citep{openai2024learning, deepseekai2025deepseekr1incentivizingreasoningcapability}
have further amplified the benefits of test-time computation by enabling multi-step deliberation. \looseness=-1

\begin{figure}[t]
  \centering
  \includegraphics[width=.9\columnwidth]{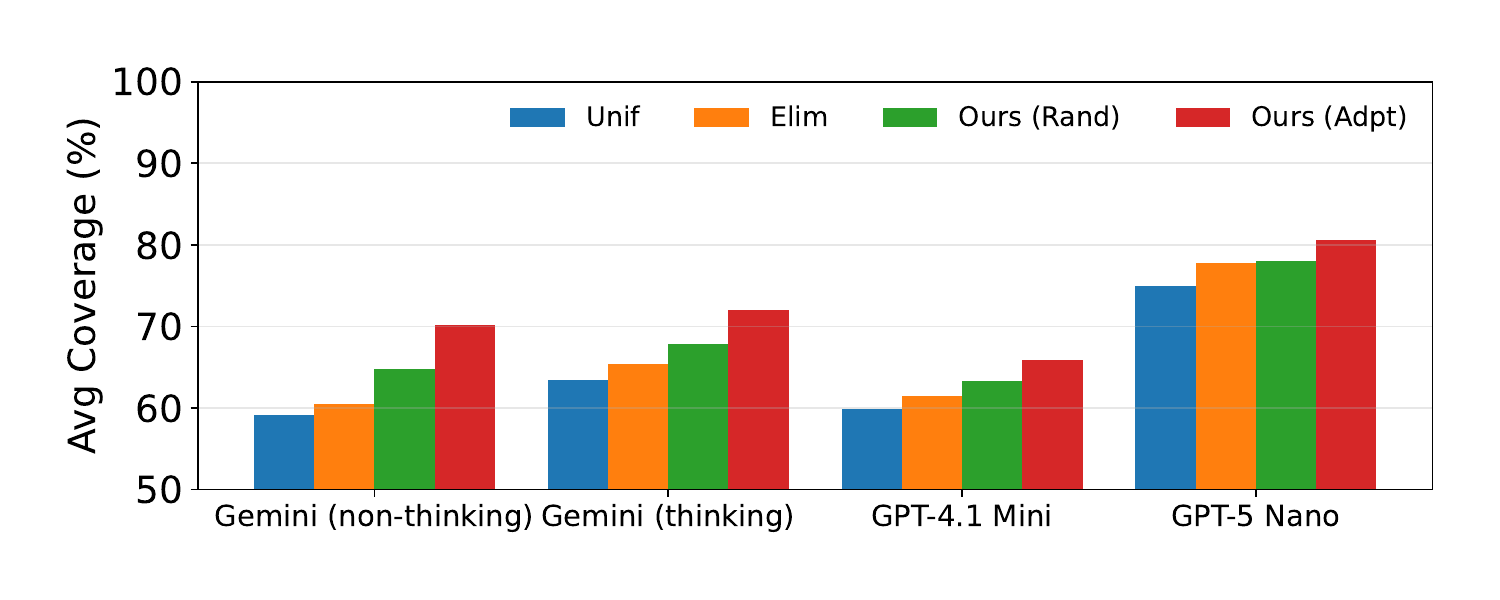}
  \caption{Average coverage (\%) across 4 model variants, where each result is averaged over 5 benchmarks spanning math, coding, and reasoning. See \cref{sec:experiments} for details.}
  \label{fig:gemini_hist}
\end{figure}

Despite this progress, most existing test-time scaling approaches share a common limitation: they rely on a largely static generation process and allocate samples uniformly across test questions. Representative examples include Best-of-$N$ methods \citep{snell2025scaling, brown2024largelanguagemonkeysscaling}. While such approaches are simple to implement and admit favorable theoretical guarantees, their lack of adaptivity makes them suboptimal across heterogeneous datasets. A more flexible line of work extends this paradigm by allocating a question-specific number of samples, typically based on estimated question difficulty \citep{wang-etal-2025-make, damani2024learning, zuo2025strategicscalingtesttimecompute}. 
Even in these methods, however, the generation distribution for each question remains static and fixed throughout the whole generation process.
A separate line of work instead modifies the sampling behavior itself \citep{liu2025can, 10.1007/s10489-025-07044-6, wu2025roletemperaturesamplingtesttime, tuyls2026representationbased}, but these methods typically require running a dedicated search procedure for each individual query or auxiliary exploration mechanisms, adding inference-time overhead and complicating deployment.
\looseness=-1

In this work, we propose a unified view of \emph{adaptive test-time compute allocation}
that jointly considers \emph{how much compute to allocate} and \emph{how that compute is used to improve the sampling distribution} at inference. Our key insight is that effective test-time scaling should not only decide \emph{where} to spend computation, but also \emph{how} additional computation influences the model’s conditional generation behavior.
We achieve this through self-adaptive in-context learning (ICL), which reuses
previously generated responses as contextual demonstrations, allowing subsequent
generations to adapt in a lightweight manner without complicated optimization procedures. \looseness=-1

Concretely, our method proceeds in two stages. A lightweight warm-up phase allocates a small, fixed budget to every query, resolving easy cases early and assembling an initial pool of question-response pairs from the test set itself. An adaptive phase then concentrates further computation on unresolved queries and, rather than resampling from the base distribution, conditions each generation on demonstrations retrieved from the evolving pool via semantic similarity. As more queries are resolved, the pool grows and the effective sampling distribution shifts---unifying compute allocation and distributional adaptation within a single inference-time loop.
\looseness=-1
Empirically, our method yields consistent improvements across a range of challenging reasoning benchmarks, including MATH-500 \citep{lightman2024lets}, GPQA-Diamond \citep{rein2024gpqa}, LiveCodeBench \citep{jain2025livecodebench}, and MinervaMath \citep{10.5555/3600270.3600548}.
Across multiple random seeds and evaluation rounds, our approach consistently outperforms a uniform allocation baseline and an adaptive allocation baseline, achieving higher performance at lower compute budgets. Notably, the gains emerge early during test-time scaling, indicating more efficient allocation of computation rather than reliance on late-stage oversampling. 
\looseness=-1

\section{Related Work}
\label{sec:related}
\paragraph{Test-time scaling.}
A growing line of research investigates how additional computation at inference time can improve the performance of large language models, without modifying their parameters.
This paradigm---often referred to as test-time compute or test-time scaling---has shown that the use of inference-time resources can close or even surpass the gap between prompting-based methods and finetuning
\citep{NEURIPS2020_1457c0d6, mosbach-etal-2023-shot}, and that encouraging longer deliberation or “thinking” at inference time yields consistent gains on reasoning benchmarks
\citep{muennighoff2025s1, deepseekai2025deepseekr1incentivizingreasoningcapability}.
Among these approaches, repeated sampling has proven particularly effective:
methods such as Best-of-$N$ generate multiple candidate responses and rely on reward models or verifiers to identify high-quality outputs
\citep{brown2024largelanguagemonkeysscaling, snell2025scaling, cobbe2021trainingverifierssolvemath, lightman2024lets, liu2025can}.
Subsequent variants further modulate the generation process based on intermediate signals such as model confidence or partial correctness
\citep{sun2024fast, manvi2024adaptiveinferencetimecomputellms, tan2025adaptiverectificationsamplingtesttime}.
While effective, these methods typically operate at the level of individual queries, adjusting inference effort locally without coordinating computation across inputs.

\paragraph{Test-time compute allocation.}
A separate but related direction considers how test-time computation should be distributed across an entire set of queries,
avoiding over-investment in easy instances while reserving resources for harder ones.
While test-time compute allocation has been demonstrated effective, most existing work rely on staged pipelines, auxiliary predictors, or pre-computed signals to guide allocation decisions \citep{damani2024learning, wang-etal-2025-make}.
One recent work \citep{zuo2025strategicscalingtesttimecompute} formalizes test-time compute allocation as pure-exploration-style bandit learning \citep{bubeck2009pure, jamieson2014best, locatelli2016optimalalgorithmthresholdingbandit, zhu2020robust, zhu2021pure, zhu2022near}, which enables fully adaptive compute allocation without substantial overhead.
However, to our knowledge, all existing allocation methods \citep{damani2024learning, wang-etal-2025-make, zuo2025strategicscalingtesttimecompute} sample from a static generation distribution for each question and are not adaptive in terms of \emph{how} individual generation is performed.
Our work builds on this line of research by  coordinating computation across queries while also allowing inference-time generation behavior to evolve during scaling.
\looseness=-1

\paragraph{In-context learning.}
In-context learning (ICL) enables language models to adapt their behavior at inference time by conditioning on demonstrations provided in the prompt, without any parameter updates \citep{NEURIPS2020_1457c0d6, agarwal2024manyshot, bertsch-etal-2025-context}. Demonstrations are typically selected via similarity-based retrieval from a labeled set \citep{liu2021makesgoodincontextexamples, xu2023knn, tanwar-etal-2023-multilingual}, and a consistent finding is that semantically relevant examples yield more reliable performance gains \citep{yoran2024making}. Beyond retrieval from fixed datasets, recent work explores constructing demonstrations dynamically from model outputs themselves \citep{madaan2023selfrefine, qin2024context, acikgoz2025selfimprovingllmagentstesttime}. 
\citet{xia2025rethinkingunsolvableincontextsearch} study ICL specifically in the test-time scaling regime, but focus on individual queries in isolation.
In this paper, we leverage ICL as a lightweight mechanism for reshaping the generation distribution within an adaptive test-time compute allocation framework---using successfully solved test queries as evolving demonstrations to guide inference on remaining ones.
\looseness=-1

\section{Problem Setting}
\label{sec:setting}

We study how to adaptively allocate inference-time compute across test questions based on their estimated difficulty, with the goal of improving performance over static compute allocation schemes that ignore variation in question difficulty.

We begin with the simplest setting, in which test-time computation is used solely for response generation.
Given a test question $x \in \cX$, the language model defines a fixed conditional distribution $p(\cdot \mid x)$.
Applying additional computation to $x$ corresponds to drawing additional samples from this distribution. Formally, let $c(x_i) \in \mathbb{N}$ denote the number of responses generated for test question $x_i$ up to a given point in the inference process. The corresponding response set is 
\[
g(x_i; c(x_i)) = \{y_1, \dots, y_{c(x_i)}\},  y_j \sim p(\cdot \mid x_i).
\]

Rather than fixing $c(x_i)$ in advance, adaptive allocation determines these values \emph{progressively}, based on intermediate feedback from previously generated responses.
The goal is to concentrate computation on unresolved or difficult questions while avoiding unnecessary sampling for questions that can be answered confidently with fewer generations.\looseness=-1

We evaluate performance using \emph{coverage} \citep{brown2024largelanguagemonkeysscaling, zuo2025strategicscalingtesttimecompute}.
A test question is considered correctly answered if at least one generated response is correct:
\[
\begin{aligned}
\coveragemath(x_i; c(x_i))
= \mathbb{I}\Bigl\{\exists\, y \in g(x_i; c(x_i)) : y \text{ correctly answers } x_i \Bigr\}.
\end{aligned}
\]

From this perspective, test-time compute allocation can be viewed as a sequential decision process that repeatedly selects which questions to allocate additional computation to, until all questions are resolved or further computation becomes unproductive.
Uniform strategies that allocate the same number of samples to every question ignore differences in difficulty and often waste computation.
In contrast, adaptive allocation aims to resolve easy questions early and reserve additional computation for harder cases, achieving strong performance with substantially lower overall compute.

\section{Methods}
\label{sec:methods}
\subsection{Intuition}

\begin{table}[t]
\caption{Coverage (\%) vs.\ number of samples per question.}
\centering
\small
\setlength{\tabcolsep}{5pt}
\renewcommand{\arraystretch}{1.15}
\begin{tabular}{lcccccccc}
\toprule
\textbf{\#Samples} & 1 & 2 & 3 & 4 & 5 & 6 & 7 & 8 \\
\midrule
Fixed     & 65.2 & 72.7 & 77.8 & 79.8 & 81.3 & 82.3 & 83.8 & 85.9 \\
Neighbour & 65.2 & 75.8 & 80.3 & 81.3 & 82.8 & 85.9 & 86.4 & 87.9 \\
\bottomrule
\end{tabular}
\label{tab:fixed-vs-dynamic}
\end{table}

Recent studies suggest that increasing diversity in generated responses is an important factor in effective test-time scaling
\citep{10.1007/s10489-025-07044-6, wu2025roletemperaturesamplingtesttime}.
We view this effect as a consequence of how test-time interventions reshape the model’s sampling distribution. However, most existing adaptive sampling methods focus primarily on \emph{where} to allocate compute—e.g., concentrating samples on harder queries—while continuing to draw samples from a fixed conditional distribution. As a result, additional sampling often yields diminishing returns when correct solutions have low probability under the base distribution.

One natural approach to address this limitation is to modify the sampling distribution itself. In-context learning (ICL) provides a flexible mechanism for inducing such shifts by conditioning generation on additional examples. However, prompting is not always beneficial: added context can degrade performance due to context-length effects, demonstration bias, or poorly matched examples \citep{yoran2024making, du-etal-2025-context}. A consistent finding in the ICL literature is that similarity-based demonstration selection improves reliability. Methods such as kNN prompting and retrieval-augmented ICL show that conditioning on semantically related examples reduces noise and leads to more stable performance gains \citep{liu2021makesgoodincontextexamples, tanwar-etal-2023-multilingual, xu2023knn, yoran2024making}.

Taken together, these findings suggest that the effectiveness of test-time interventions depends not only on how compute is allocated, but also on whether the induced distribution shift is well aligned with the target query. To illustrate this effect, we conduct a controlled experiment on the GPQA-Diamond dataset, where each question is evaluated under an identical sampling budget using either a fixed prompt or a prompt constructed from semantically neighboring questions. As shown in \cref{tab:fixed-vs-dynamic}, neighbor-based prompting consistently yields higher coverage, indicating that similarity-guided distribution shifts enable more effective use of test-time compute.

\subsection{Our algorithm}
Given a set of test questions $\cS = \{x_1, \dots, x_n\}$, our goal is to improve test-time performance by applying an adaptive scaling strategy. Our approach leverages prior test-time scaling methods and retrieval-based methods by continuously reusing generated and evaluated test-time samples to update in-context prompts, enabling the conditional response distribution to evolve over time rather than remain fixed.
This evolving mechanism operates entirely at inference time, without modifying model parameters. 

Our method proceeds in two stages:
\begin{enumerate}
    \item A \emph{Warm-up stage}, which applies a lightweight, fixed distribution sampling strategy to identify easy queries and construct an initial $D_{test}$ of candidate demonstrations.
    \item An \emph{Adaptive Distribution Sampling stage}, which focuses more computation on unresolved queries and conditions future generations on evolving $D_{test}$.
\end{enumerate}

This two-stage design enables the model to concentrate inference on challenging queries while actively reshaping the response distribution through evolving in-context demonstrations. The full procedure is summarized in \cref{alg:methods}. The framework is modular and can accommodate alternative test-time sampling interventions, demonstration selection strategies, and aggregation rules. 

\subsubsection{Warm-up}
\label{subsec:warmup}

In the warm-up stage, we allocate a small, fixed amount of compute to each query.
For each $x \in \cS$, we generate a fixed number of responses
\[
g_0(x) = \{y_1, \dots, y_K\}, \qquad y_i \sim p(\cdot \mid x).
\]
Each response is evaluated by a reward oracle $r(x,y)$ usually with a range $[0,1]$.
If any response satisfies $r(x,y) \ge \gamma$, then the question is marked as solved and removed from further consideration. Given that our evaluation metric is \emph{coverage}, we use ground-truth as our reward model $r$ to evaluate the correctness directly for this step and a $\gamma = 1$ correspondingly. \looseness=-1 The warm-up stage serves two purposes. First, it resolves easy queries early, preventing unnecessary compute expenditure, including ICL-construction, which increases input token usage. Second, it constructs an initial $D_{test}$ that will later serve as candidates for ICL demonstrations. This stage also naturally serves as a difficulty estimation process over the test set.

\begin{figure*}[!tb]
  \centering
  \includegraphics[width=\linewidth]{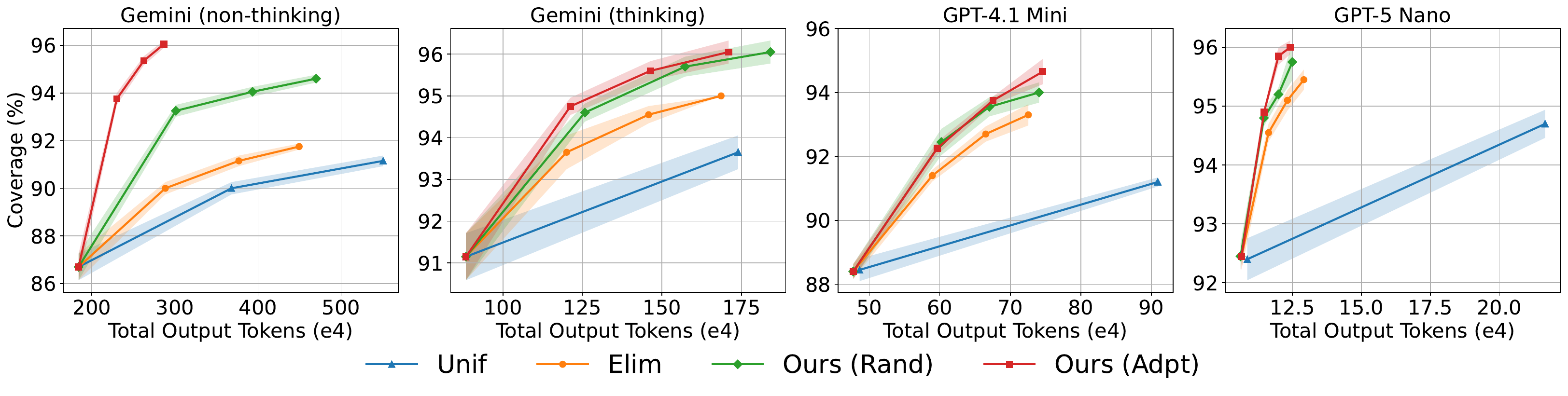}
  \caption{Coverage comparison on MATH-500.}
  \label{fig:math500}
\end{figure*}

\subsubsection{Adaptive Distribution Sampling}
\label{subsec:adaptive}

After the warm-up stage, the remaining test questions correspond to instances for which correct responses have low probability under the model’s base conditional distribution. As a result, additional independent sampling from the same distribution is unlikely to yield success.

Unlike prior scaling methods \citep{brown2024largelanguagemonkeysscaling,  zuo2025strategicscalingtesttimecompute, wang-etal-2025-make}, which repeatedly sample from a fixed distribution $p(\cdot \mid x)$, our approach modifies the sampling distribution itself by conditioning on ICL demonstrations that draw from the model itself. However, in terms of the pure resource allocation aspect, we adopt an effective and strong adaptive allocation strategy that focuses compute on harder queries as easier ones are resolved. Building on this strong foundation, we further improve efficiency by adapting the sampling distribution via evolving ICL conditioning, rather than sampling repeatedly from a fixed distribution. We retain the strengths of adaptive resource allocation and further enhance it by adaptively shaping the generation distribution.

\paragraph{ICL construction via similarity-based selection.}
For each active query $x \in \cA$, we construct an ICL prompt by selecting a set of demonstrations from $D_{test}$.
We first embed each question using a fixed embedding function $\phi(\cdot)$ and measure semantic similarity using cosine similarity.
The neighborhood of $x$ is defined as the set of top-$P$ most similar queries:
\[
\begin{aligned}
\mathcal{P}(x)
&= \operatorname{Top}\text{-}P\Big(\cA \setminus \{x\},\;
\cos\big(\phi(x), \phi(\cdot)\big)\Big),
\end{aligned}
\]
where $\text{cos}(\cdot, \cdot )$ denotes the cosine similarity.
For each neighbor question $z \in \mathcal{P}(x)$, let $\tilde y(z)$ denote the most recently generated response for $z$.
In practice, since evaluation is deterministic and queries are eliminated once a correct response is found, $\tilde y(z)$ is typically the first correct response for $z$ when available.
The ICL prompt for $x$ is then constructed as
\[
\Pi(x) = \{(z, \tilde y(z)) \mid z \in \mathcal{P}(x)\}.
\]
This prompt is used to condition subsequent generations for $x$.

\paragraph{ICL-conditioned sampling.}
Given the constructed prompt $\Pi(x)$, new responses are generated from the conditional distribution
\[
y \sim p(\cdot \mid x, \Pi(x)).
\]
These responses are still evaluated by the reward oracle, and the active set $A$ is updated accordingly.
\begin{algorithm}[H]
\caption{Self-Evolving Test-Time Allocation}
\label{alg:methods}
\begin{algorithmic}[1]
\renewcommand{\algorithmicrequire}{\textbf{Require:}}
\renewcommand{\algorithmicensure}{\textbf{Return:}}

\REQUIRE Test questions $\cS$, language model $p$, reward oracle $r$, total generation rounds $R$, warm-up rounds $R_{\text{warm}}$ ($R_{\text{warm}} \le R$), per-round sample size $K$, elimination threshold $\gamma$, embedding function $\phi(\cdot)$, neighborhood size $P$.
\ENSURE For each $x \in \cS$, a set of generated samples $g(x)$ and a final answer $\tilde y(x)$.

\STATE Initialize response pool $g(x) \gets \emptyset$ for all $x \in \cS$ \algcommentlight{stores all test-time generations}
\STATE Initialize $\tilde y(x) \gets \emptyset$ for all $x \in \cS$ 
\STATE Initialize active set $\cA \gets \cS$ 

\vspace{0.35em}
\STATE \textcolor{blue}{\textbf{Stage 1: Warm-up (Difficulty Estimation and $\cD_{\text{test}}$ Construction)}} \algcommentlight{fixed distribution sampling}

\FOR{round $t = 1$ to $R_{\text{warm}}$}
    \FOR{each $x \in \cS$}
        \STATE Generate $K$ responses $\{y_i\}_{i=1}^K \sim p(\cdot \mid x)$ \algcommentlight{base distribution}
        \STATE Append responses to $g(x)$ 
        \IF{$\exists\, y \in g(x)$ such that $r(x,y) \ge \gamma$}
            \STATE $\tilde y(x) \gets y$ \algcommentlight{first correct (deterministic verifier)}
            \STATE $\cA \gets \cA \setminus \{x\}$ \algcommentlight{eliminate solved questions from the active set}
        \ENDIF
    \ENDFOR
\ENDFOR

\vspace{0.35em}
\STATE \textcolor{orange}{\textbf{Stage 2: Adaptive Distribution Sampling}} \algcommentlight{self-improving ICL}

\STATE Precompute question embeddings $\phi(x)$ for all $x \in \cS$ 
\FOR{each $x \in \cS$}
    \STATE Define neighborhood:
    \[
    \mathcal{P}(x) \gets \operatorname{Top}\text{-}P\big(\cS \setminus \{x\},\;
    \mathrm{cos}(\phi(x), \phi(\cdot))\big)
    \]
    \algcommentlight{fixed neighbors; prompts evolve via $\tilde y(\cdot)$}
\ENDFOR

\FOR{round $t = R_{\text{warm}}+1$ to $R$}
    \FOR{each $x \in \cA$} 
        \STATE Construct ICL prompt using most recent neighbor responses:
        \[
        \Pi(x) \gets \{(z, \tilde y(z)) : z \in \mathcal{P}(x),\; \tilde y(z)\ \text{exists}\}
        \]
        \algcommentlight{updates as $\cD_{\text{test}}$ grows}
        \STATE Generate $K$ responses $\{y_i\}_{i=1}^K \sim p(\cdot \mid x, \Pi(x))$ \algcommentlight{distribution shift via ICL}
        \STATE Append responses to $g(x)$
        \IF{$\exists\, y \in g(x)$ such that $r(x,y) \ge \gamma$}
            \STATE $\tilde y(x) \gets y$ 
            \STATE $\cA \gets \cA \setminus \{x\}$ \algcommentlight{eliminate upon success}
        \ENDIF
    \ENDFOR
\ENDFOR

\RETURN $\{g(x), \tilde y(x)\}_{x \in \cS}$ 

\end{algorithmic}
\end{algorithm}

\begin{figure*}[!tb]
  \centering
  \includegraphics[width=\linewidth]{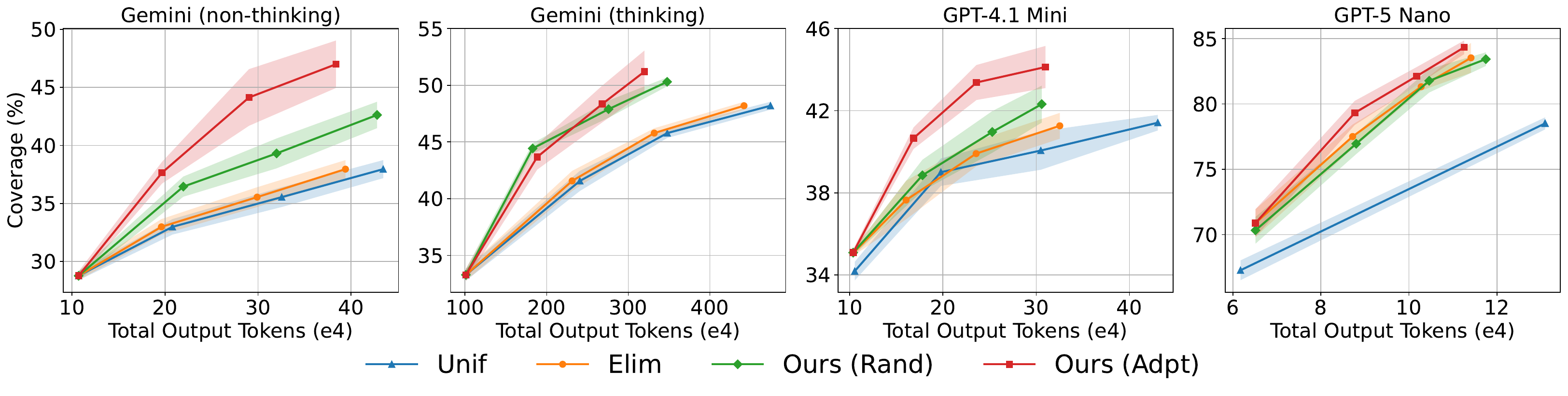}
  \caption{Coverage comparison on LiveCodeBench.}
  \label{fig:lcb}
\end{figure*}

\begin{figure*}[t]
  \centering
  \includegraphics[width=\linewidth]{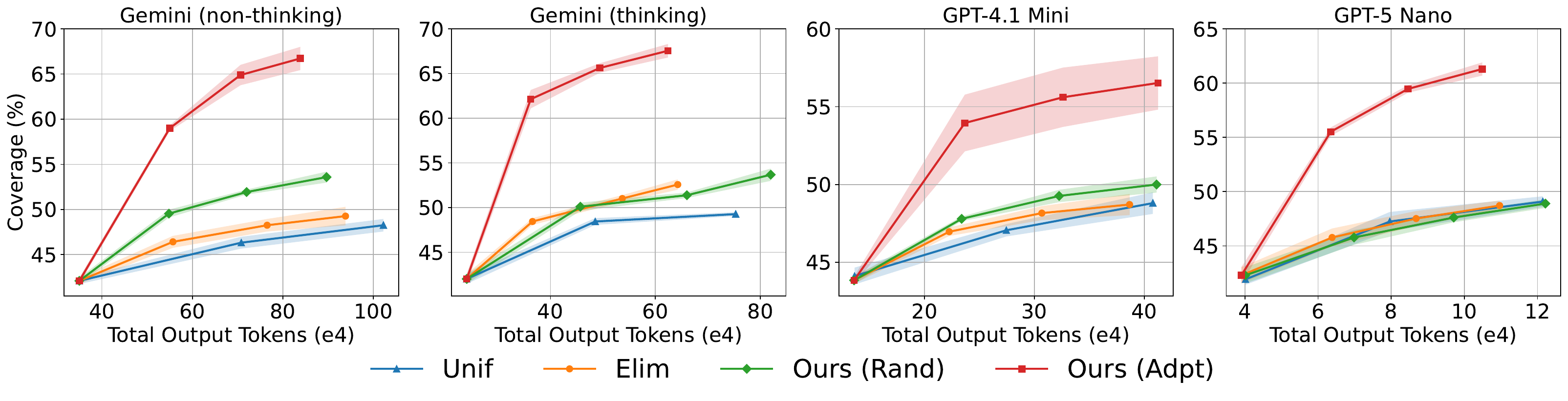}
  \caption{Coverage comparison on MinervaMath.}
  \label{fig:minerva}
\end{figure*}

\section{Experiments}
\label{sec:experiments}
In this section, we present the evaluation of our approach among different datasets with different models. 

\subsection{Experimental settings}

\paragraph{Datasets and models.}
We evaluate our method on four widely used reasoning and coding benchmarks: 
MATH-500 \citep{lightman2024lets}, 
LiveCodeBench \citep{jain2025livecodebench}, 
MinervaMath \citep{10.5555/3600270.3600548}, and 
GPQA-Diamond \citep{rein2024gpqa}. 
For LiveCodeBench, we restrict our evaluation to questions released between 10/05/2024 and 01/04/2025, following the evaluation protocol reported for Gemini~2.5 \citep{comanici2025gemini25pushingfrontier}.  

In addition to these benchmarks, we construct a custom evaluation dataset using Reasoning Gym \citep{stojanovski2025reasoning}. 
This dataset consists of 400 math-oriented reasoning problems sampled at the \emph{hard} difficulty level. 
Details of the dataset construction process are provided in \cref{app:experiments}. For MATH-500 datasets, there are 500 math reasoning questions. There are 272 reasoning questions in the MinervaMath dataset, and 198 reasoning questions in GPQA-Diamond correspondingly. The selected LiveCodeBench dataset contains 166 test questions relating to coding. 

We evaluate both reasoning and non-reasoning large language models drawn from recent frontier model families. 
Specifically, we use the  \texttt{Gemini-2.5-flash-lite} model under two configurations: with internal thinking enabled and disabled.
We also include \texttt{GPT-4.1 Mini}, a non-reasoning model, and \texttt{GPT-5 Nano}, a reasoning-oriented model. 
All models are accessed exclusively through their respective public APIs.

\begin{figure*}[t]
  \centering
  \includegraphics[width=\linewidth]{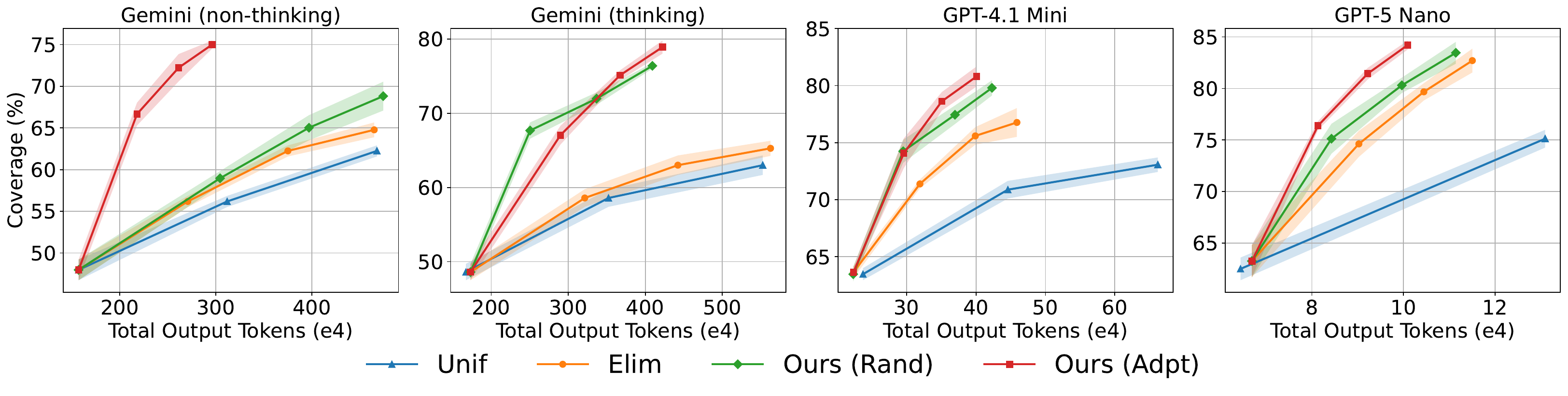}
  \caption{Coverage comparison on GPQA-Diamond.}
  \label{fig:gpqa}
\end{figure*}

\paragraph{Baselines and metrics.}
We compare our method \textsc{Ours (Adpt)} against three methods.
The first is the Best-of-$N$ (\textsc{Unif}) strategy \citep{snell2025scaling}, which uniformly allocates test-time compute by sampling $N$ independent responses per query from a fixed conditional distribution and selecting the highest-scoring one using a reward oracle. The second method is the \textsc{Elimination} (\textsc{Elim}) method \citep{manvi2024adaptiveinferencetimecomputellms, zuo2025strategicscalingtesttimecompute}, which formulates test-time scaling as an adaptive allocation problem.
In this method, test questions that achieve a correct response early are eliminated from further sampling, while unresolved questions continue to receive additional samples. All samples for a given query are drawn from a fixed conditional distribution, and adaptation occurs solely through the allocation of the number of samples generated per question. The last method \textsc{Ours (Rand)} is a variant of our approach that uses \emph{random} prompt selection. This method follows the same two-stage procedure as our method but replaces similarity-based demonstration selection with randomly chosen test-time examples.

\begin{figure*}[t]
  \centering
  \includegraphics[width=\linewidth]{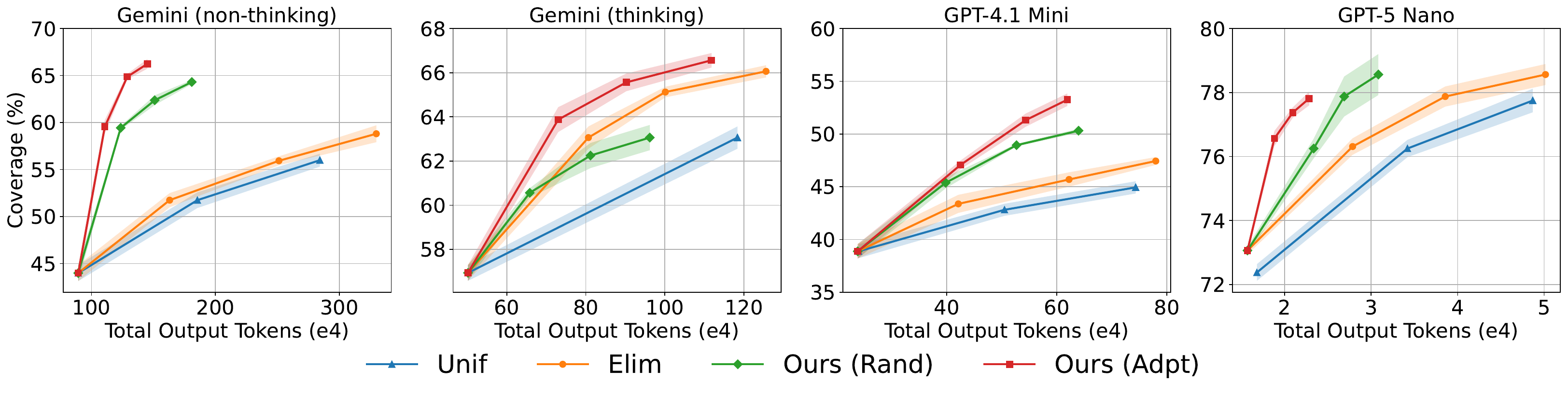}
  \caption{Coverage comparison on Reasoning Gym.}
  \label{fig:rg}
\end{figure*}

All results are averaged over four random runs, with shaded regions in plots indicating $\pm 0.5$ standard deviations. 
As stated in \cref{alg:methods}, we adopt a round-based evaluation for all methods, where in each round, new responses are generated \textbf{only for test questions that remain unsolved}. 
By default, we use a total of $R = 4$ rounds and $R_{warm} = 1$. For the number of neighboring shots in adaptive rounds, we apply $P=3$ for all our settings. For \textsc{Unif}, \textsc{Elim}, and the warm-up round, we apply zero-shot COT prompt \citep{10.5555/3600270.3601883}. We include all details in \cref{app:additional_experiments}.
For \textsc{unif}, we match the same level of output token to that of \textsc{Elim} to ensure a fair comparison. We report performance curves for all methods using the same random seeds. For \textsc{Elim}, \textsc{Ours (Rand)}, and \textsc{Ours (Adpt)}, we use the same warm-up stage generation for efficiency purposes, which corresponds to the first point in each plot.
For Gemini-family models, results are fully reproducible under identical seeds, and therefore, the first-point performance is exactly matched across methods.
For GPT-based models, perfect seed replication is not supported, and as a result, the first-point performance may exhibit minor variation despite using the same seed.
As stated before, we primarily report \emph{coverage} as our main evaluation metric because the \emph{accuracy} is highly relevant to the \emph{coverage} performance. Nevertheless, we provide the \emph{accuracy} performance to complement our selection metrics in \cref{sec:ablation}. 

\subsection{Overall experiment results}

\paragraph{Benchmark results.} \cref{fig:allhist} reports the final round \emph{coverage} of all methods across all benchmarks and models. For the \textsc{Unif} baseline, we propagate the current final-round performance to unfinished rounds (till $R$ at 4) to enable a fair comparison under the same evaluation protocol.

Across all benchmarks and model families, our method consistently improves coverage relative to \textsc{Unif}, demonstrating the benefit of adaptive test-time scaling over uniform sampling.
Compared to \textsc{Elim}, our approach yields additional gains on nearly all benchmarks, highlighting the advantage of distributional adaptation via evolving in-context demonstrations.
The only exception occurs for \texttt{GPT-5 Nano} when compared against \textsc{Elim} and \textsc{Ours (Rand)} on Reasoning Gym. This behavior is consistent with the relatively limited sensitivity of  \texttt{GPT-5 Nano} to prompt-based conditioning on Reasoning Gym.
\looseness=-1

While random prompts can yield improvements in many settings, their performance is substantially less consistent.
On reasoning-intensive benchmarks such as Reasoning Gym, MinervaMath, and LiveCodeBench, \textsc{Ours (Rand)} occasionally underperforms \textsc{Unif} or \textsc{Elim}, with the effect being more pronounced for reasoning-oriented models. In contrast, similarity-based demonstration selection consistently achieves larger and more stable gains across models and benchmarks, highlighting the importance of structured prompt construction for inducing beneficial distribution shifts during test-time scaling.
Overall, these results demonstrate that our approach leads to robust and consistent coverage gains beyond some prior methods.
\looseness=-1

\paragraph{Output token efficiency.}
\label{para:efficiency}
In addition to coverage, we report total output token usage alongside performance curves in \cref{fig:math500,fig:lcb,fig:minerva,fig:gpqa,fig:rg}. We measure output token usage excluding the thinking summary for all reasoning models, as this summary is a simulated artifact and does not reflect the model’s actual internal reasoning process for any of the evaluated models. Still, we include a representative result of total token usage in \cref{app:additional_experiments}.
Across all benchmarks, our method consistently achieves the highest token efficiency among all compared approaches, for it requires fewer tokens to reach the same accuracy level.
In particular, our approach typically exceeds the performance of \textsc{Elim}, \textsc{Unif}, and \textsc{Ours (Rand)} while consuming markedly fewer samples.
These results demonstrate that our evolving distributional adaptation not only improves coverage but also yields strong gains in token efficiency across diverse tasks and models. We include an analysis of the reasons why we can achieve such token efficiency in \cref{app:token}.

\subsection{Ablation Studies}
\label{sec:ablation}

\begin{figure*}[!t]
  \centering
  \begin{subfigure}[t]{0.32\linewidth}
    \centering
    \includegraphics[width=\linewidth]{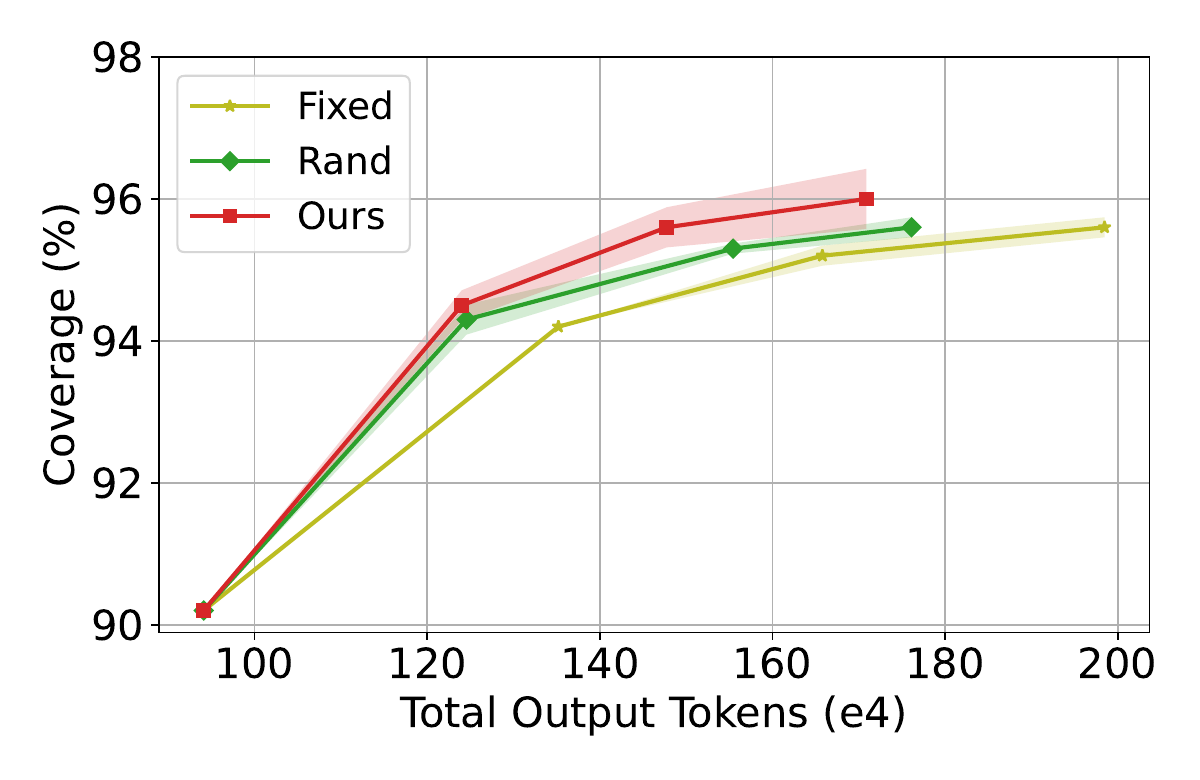}
    \caption{Fixed ICL (Gemini thinking)}
  \end{subfigure}\hfill
  \begin{subfigure}[t]{0.32\linewidth}
    \centering
    \includegraphics[width=\linewidth]{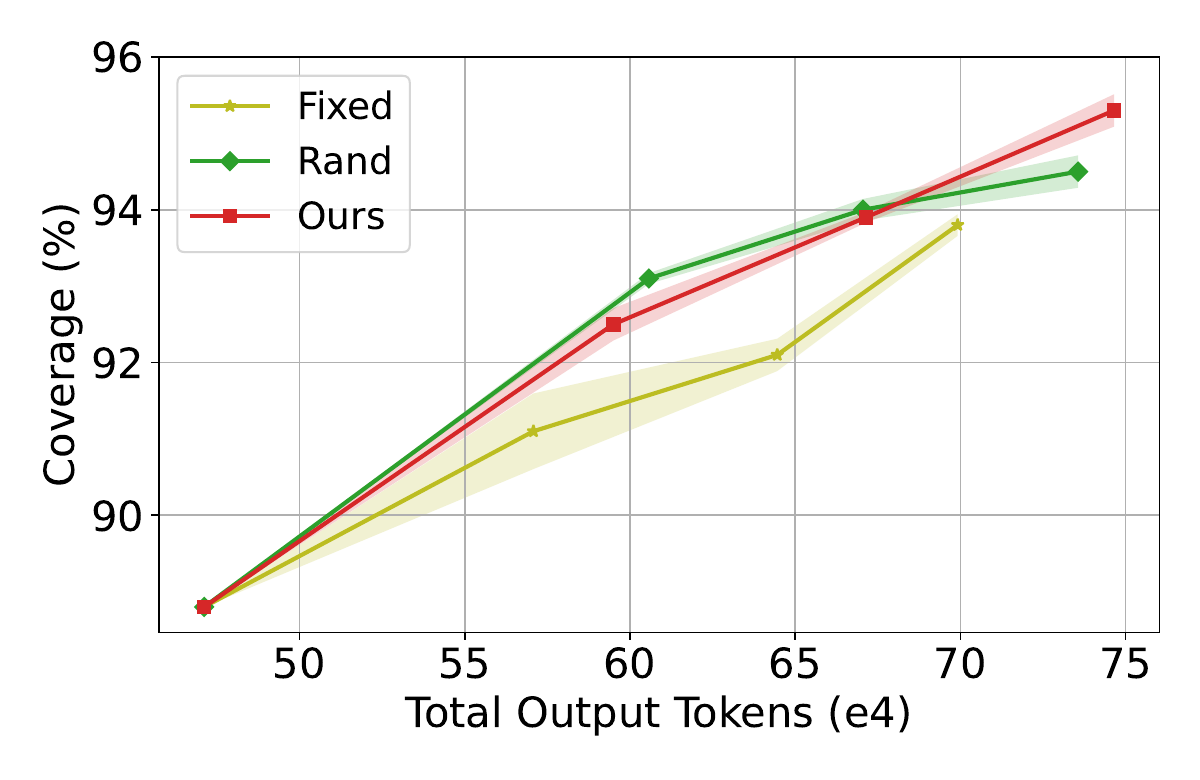}
    \caption{Fixed ICL (GPT-4.1 Mini)}
  \end{subfigure}\hfill
  \begin{subfigure}[t]{0.32\linewidth}
    \centering
    \includegraphics[width=\linewidth]{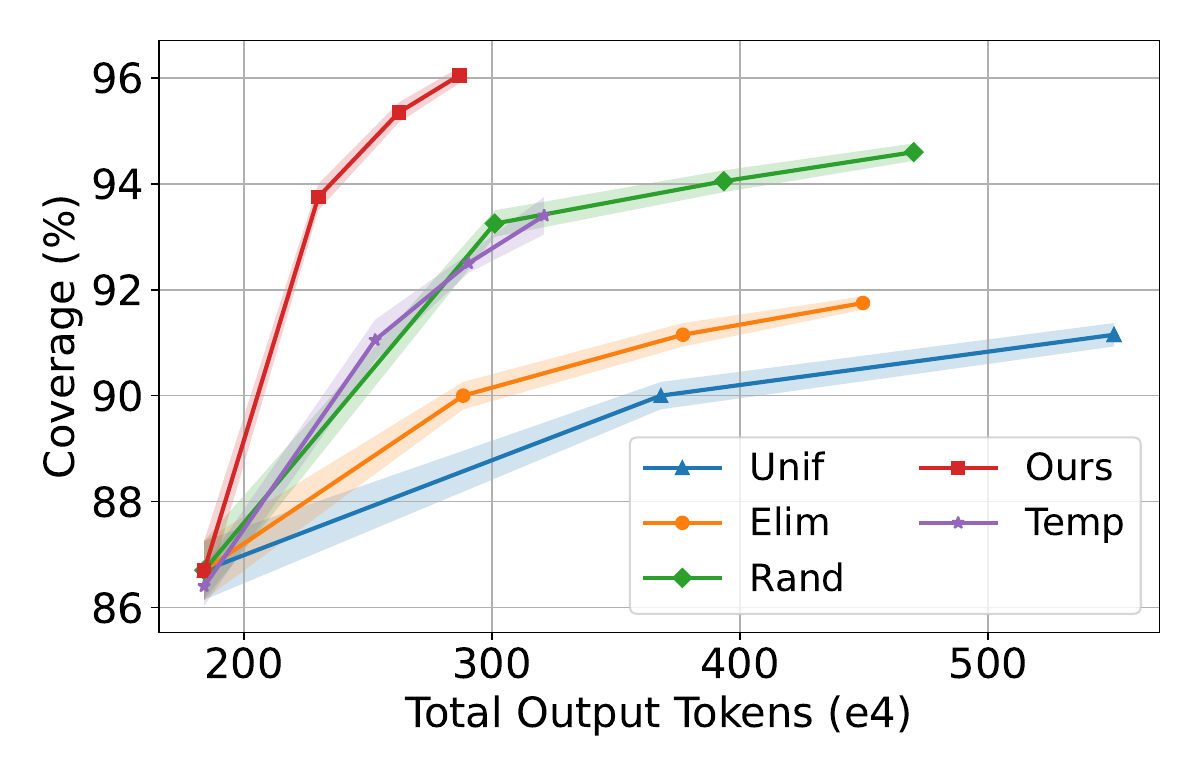}
    \caption{Temperature tuning}
  \end{subfigure}

  \vspace{0.5em}
  \begin{subfigure}[t]{0.32\linewidth}
    \centering
    \includegraphics[width=\linewidth]{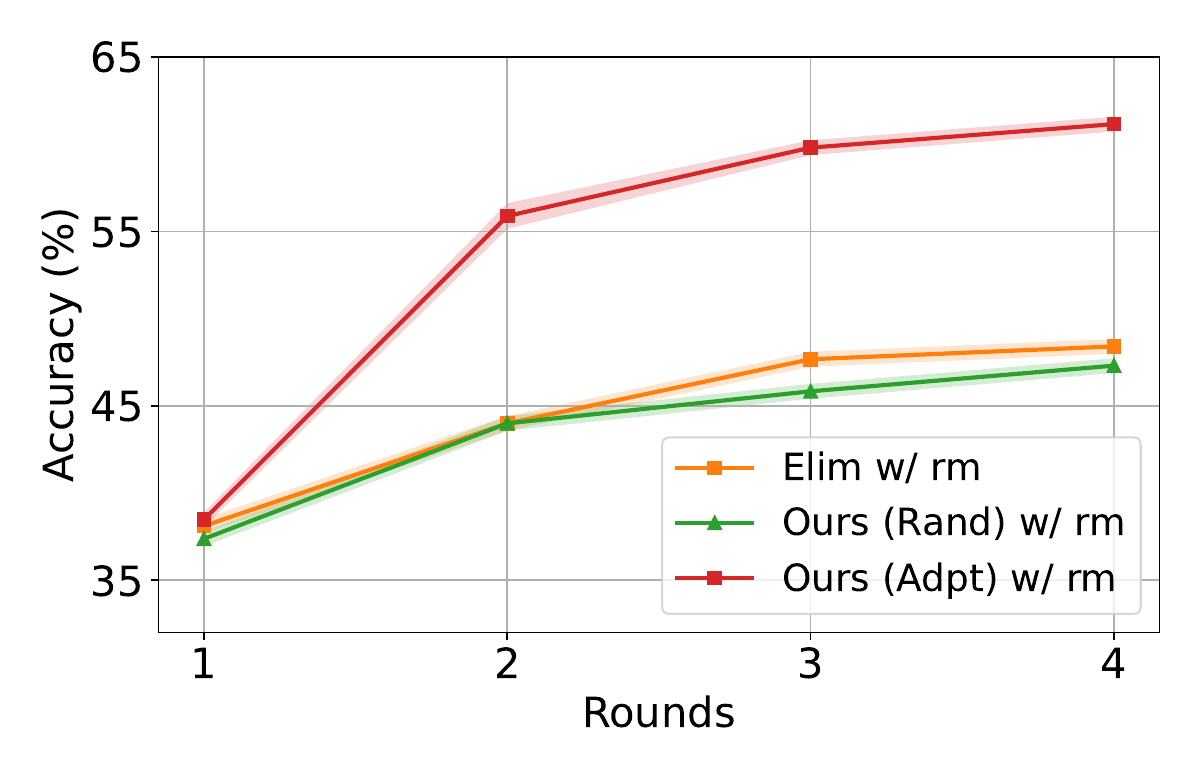}
    \caption{Reward model (Gemini thinking)}
  \end{subfigure}\hfill
  \begin{subfigure}[t]{0.32\linewidth}
    \centering
    \includegraphics[width=\linewidth]{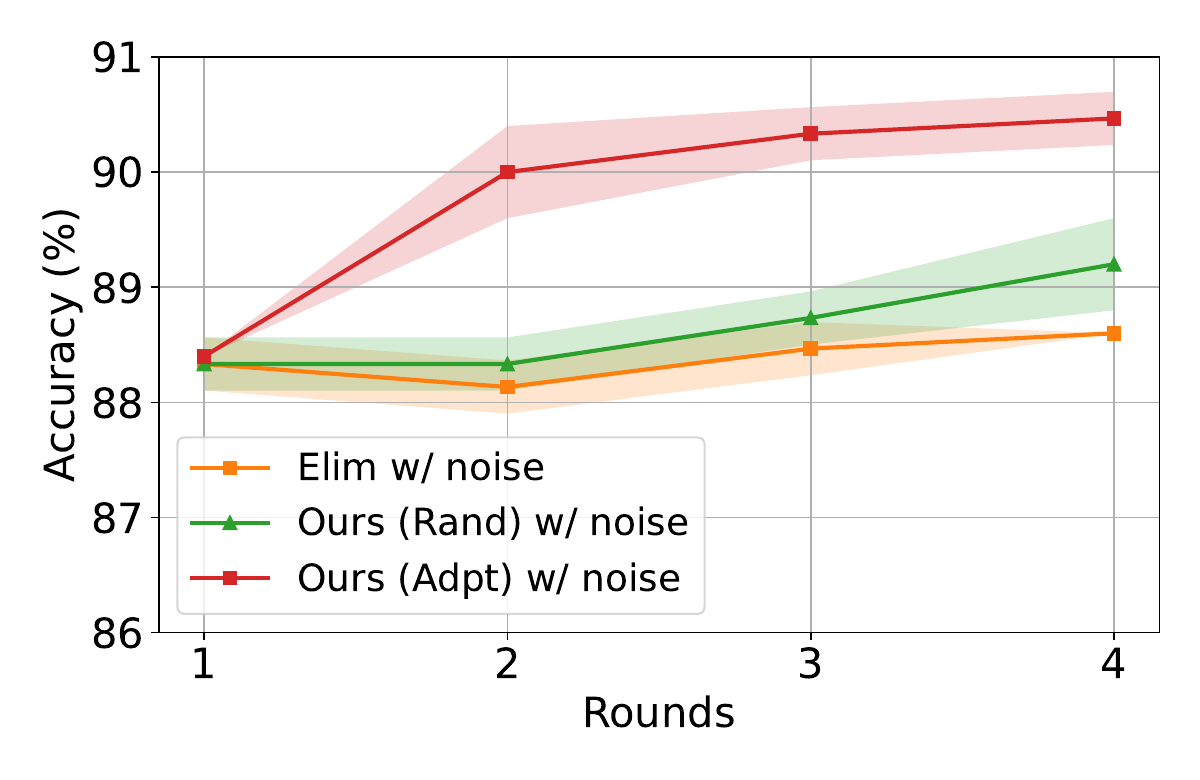}
    \caption{Noisy reward (GPT-4.1 Mini)}
  \end{subfigure}\hfill
  \begin{subfigure}[t]{0.32\linewidth}
    \centering
    \includegraphics[width=\linewidth]{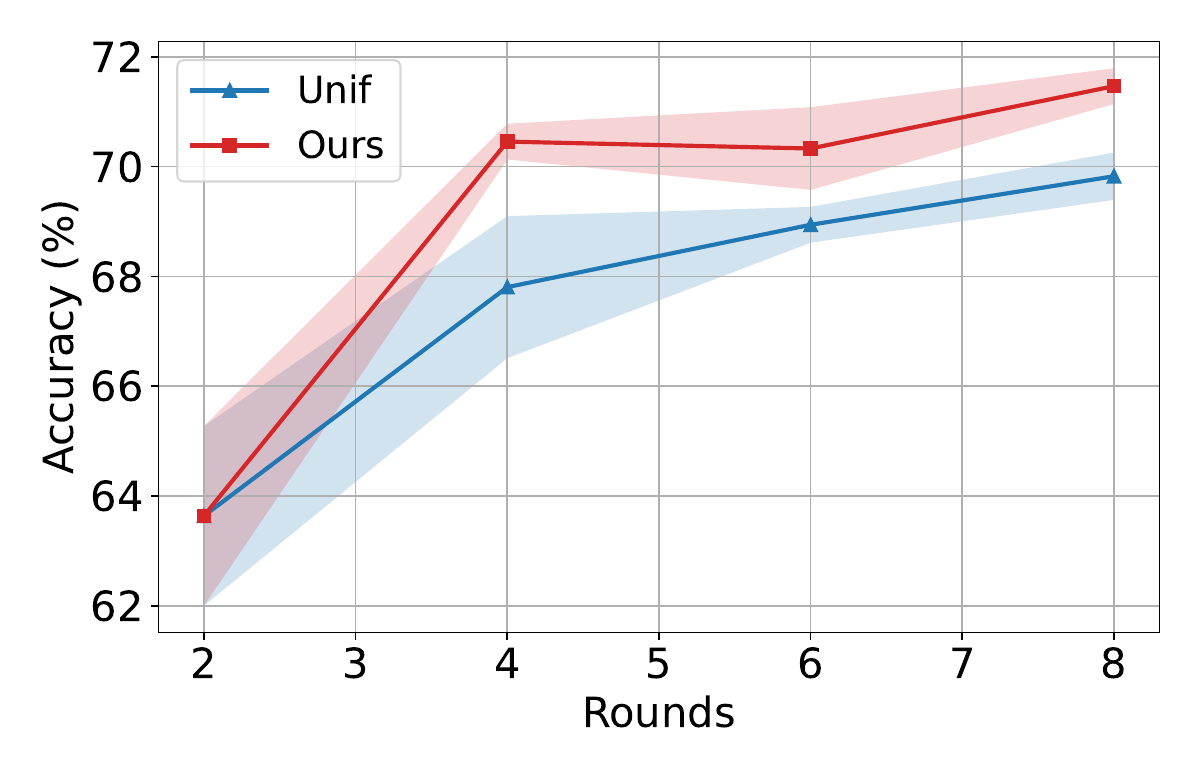}
    \caption{Self-consistency}
  \end{subfigure}

  \caption{Ablation studies across fixed demonstrations, temperature tuning, and selection strategies.}
  \label{fig:abla}
\end{figure*}
\paragraph{Fixed demonstrations.}
We further conduct an ablation study using a fixed set of in-context demonstrations with \textsc{Elim}, while keeping all other settings identical to our proposed approach.
Given the availability of demonstration prompts in prior work, we evaluate this fixed-prompt variant on the MATH-500 dataset using the \texttt{Gemini-2.5-flash-lite} model with thinking enabled and the \texttt{GPT-4 Mini} model.
The demonstrations used in this experiment follow commonly adopted prompt templates from prior studies \citep{10.5555/3600270.3600548, brown2024largelanguagemonkeysscaling}. The results are presented in \cref{fig:abla}. Our dynamic setting offers an advantage in both token efficiency and accuracy, even with random demonstration selection. Fixed demonstrations still lead to a static sampling distribution, for it does not introduce any adaptivity in the generation process. 

\paragraph{Alternative sampling distribution.}
We have highlighted the importance of the sampling distribution in test-time scaling.
Our framework is agnostic to the specific mechanism used to modify this distribution and can, in principle, accommodate any test-time intervention that reshapes model outputs.
As a simple alternative to in-context learning, we consider straightforward temperature adjustment, which directly controls the entropy of the sampling distribution.
We conduct an ablation study on the MATH-500 benchmark using the \texttt{Gemini-2.5-flash-lite} model with thinking disabled, where \cref{alg:methods} is modified to progressively increase the sampling temperature across rounds in place of updating ICL demonstrations. We start with a temperature $= 0.3$ and then increase by 0.4 every round. 
All other experimental settings are kept identical.
The results, shown in \cref{fig:abla} and marked as \textsc{Temp}, illustrate the plug-and-play nature of our framework with respect to different sampling distribution controls.

\paragraph{Self-consistency.}
 Other than the coverage metrics, we utilize self-consistency \citep{wang2023selfconsistency} as suggested in \citet{wang2025thinkdeepthinkfast}, which claims that self-consistency could have a good performance as a selection mechanism for reasoning models. We conduct experiments using \texttt{GPT-5 Nano} model on the GPQA-Diamond benchmark. We increase our rounds to $P=8$ to build a better consensus pool, and we do not eliminate based on a ground-truth oracle. Instead, we eliminate once the response pool reaches a consensus over 75\%  as a variation for our \cref{alg:methods}. Since self-consistency picks out a final answer, we report \emph{accuracy} instead of \emph{coverage}. To differentiate rounds from the previous plots, we mark rounds explicitly. Other settings are identical to our main experiments. We report our results in \cref{fig:abla}. Our method still outperforms the baselines with self-consistency selection. Also, given that our token efficiency has been shown sufficiently in other experiments, we do not explicitly show it, but our method has a better output token efficiency while achieving a better accuracy performance.

 \paragraph{Reward-model.}
We further evaluate our method under a reward-model-based setting, where answer quality is assessed using a strong external model rather than direct ground-truth comparison. Empirically, we observe that Gemini-3-flash demonstrates stronger performance in the Math category compared to the current rewardbench \citep{lambert-etal-2025-rewardbench} leader (Gemini-2.5-pro) while also offering significantly lower latency. Therefore, we adopt Gemini-3-flash as our reward model for efficiency and competitiveness. We conduct this experiment using the Gemini-2.5-flash-lite
model with thinking enabled on the MinervaMath dataset. The results in \cref{fig:abla} are consistent with our earlier findings. 

Additionally, to explicitly test robustness under imperfect verification, we introduce controlled noise into the reward model's outputs by randomly flipping binary correctness labels with a 5\% probability. In this noisy-reward setting, the reward model determines whether a response is considered solved and whether it is added to the demonstration pool; if no response is marked correct for a query, we fall back to selecting the most recent response. Ground-truth labels are used only for final evaluation and are not involved in adaptive allocation or demonstration construction. We conduct this experiment using GPT-4.1-mini on the MATH500 dataset. The results in \cref{fig:abla} still remain consistent with our earlier findings.

\section{Discussion}
\label{sec:discussion}

This work studies test-time scaling from a unified perspective that jointly considers how inference-time computation is allocated and how additional computation is used to influence the model’s sampling behavior during inference. Rather than treating test-time scaling solely as generating more samples from a fixed process, we show that allowing the conditional generation behavior to evolve over time is a critical and complementary dimension. By leveraging evolving in-context demonstrations constructed from test-time generations, our framework enables the response process to adapt progressively, without access to external training data, model updates, or auxiliary allocation models. Empirically, this structured adaptation yields both improved coverage and higher token efficiency across diverse benchmarks and model families, consistently outperforming or matching strong adaptive baselines while consuming fewer inference tokens.

More broadly, our framework highlights the importance of temporal structure in test-time inference. As generations accumulate, intermediate outputs provide increasingly informative signals that can be reused to guide subsequent computation, enabling inference-time behavior to improve even under a fixed model. While we instantiate this idea through in-context learning, the framework itself is not limited to ICL and naturally accommodates alternative mechanisms for influencing generation, such as temperature control or other lightweight decoding strategies. As inference cost becomes an increasingly central concern, we believe adaptive test-time strategies that jointly allocate computation and allow generation behavior to evolve will play a key role in the practical and efficient deployment of large language models.

\bibliography{refs}

\begin{thebibliography}{54}
\providecommand{\natexlab}[1]{#1}
\providecommand{\url}[1]{\texttt{#1}}
\expandafter\ifx\csname urlstyle\endcsname\relax
  \providecommand{\doi}[1]{doi: #1}\else
  \providecommand{\doi}{doi: \begingroup \urlstyle{rm}\Url}\fi

\bibitem[Acikgoz et~al.(2025)Acikgoz, Qian, Ji, Hakkani-Tür, and Tur]{acikgoz2025selfimprovingllmagentstesttime}
Emre~Can Acikgoz, Cheng Qian, Heng Ji, Dilek Hakkani-Tür, and Gokhan Tur.
\newblock Self-improving llm agents at test-time, 2025.
\newblock URL \url{https://arxiv.org/abs/2510.07841}.

\bibitem[Agarwal et~al.(2024)Agarwal, Singh, Zhang, Bohnet, Rosias, Chan, Zhang, Anand, Abbas, Nova, Co-Reyes, Chu, Behbahani, Faust, and Larochelle]{agarwal2024manyshot}
Rishabh Agarwal, Avi Singh, Lei~M Zhang, Bernd Bohnet, Luis Rosias, Stephanie~C.Y. Chan, Biao Zhang, Ankesh Anand, Zaheer Abbas, Azade Nova, John~D Co-Reyes, Eric Chu, Feryal Behbahani, Aleksandra Faust, and Hugo Larochelle.
\newblock Many-shot in-context learning.
\newblock In \emph{The Thirty-eighth Annual Conference on Neural Information Processing Systems}, 2024.
\newblock URL \url{https://openreview.net/forum?id=AB6XpMzvqH}.

\bibitem[Bertsch et~al.(2025)Bertsch, Ivgi, Xiao, Alon, Berant, Gormley, and Neubig]{bertsch-etal-2025-context}
Amanda Bertsch, Maor Ivgi, Emily Xiao, Uri Alon, Jonathan Berant, Matthew~R. Gormley, and Graham Neubig.
\newblock In-context learning with long-context models: An in-depth exploration.
\newblock In Luis Chiruzzo, Alan Ritter, and Lu~Wang, editors, \emph{Proceedings of the 2025 Conference of the Nations of the Americas Chapter of the Association for Computational Linguistics: Human Language Technologies (Volume 1: Long Papers)}, pages 12119--12149, Albuquerque, New Mexico, April 2025. Association for Computational Linguistics.
\newblock ISBN 979-8-89176-189-6.
\newblock \doi{10.18653/v1/2025.naacl-long.605}.
\newblock URL \url{https://aclanthology.org/2025.naacl-long.605/}.

\bibitem[Brown et~al.(2024)Brown, Juravsky, Ehrlich, Clark, Le, Ré, and Mirhoseini]{brown2024largelanguagemonkeysscaling}
Bradley Brown, Jordan Juravsky, Ryan Ehrlich, Ronald Clark, Quoc~V. Le, Christopher Ré, and Azalia Mirhoseini.
\newblock Large language monkeys: Scaling inference compute with repeated sampling, 2024.
\newblock URL \url{https://arxiv.org/abs/2407.21787}.

\bibitem[Brown et~al.(2020)Brown, Mann, Ryder, Subbiah, Kaplan, Dhariwal, Neelakantan, Shyam, Sastry, Askell, Agarwal, Herbert-Voss, Krueger, Henighan, Child, Ramesh, Ziegler, Wu, Winter, Hesse, Chen, Sigler, Litwin, Gray, Chess, Clark, Berner, McCandlish, Radford, Sutskever, and Amodei]{NEURIPS2020_1457c0d6}
Tom Brown, Benjamin Mann, Nick Ryder, Melanie Subbiah, Jared~D Kaplan, Prafulla Dhariwal, Arvind Neelakantan, Pranav Shyam, Girish Sastry, Amanda Askell, Sandhini Agarwal, Ariel Herbert-Voss, Gretchen Krueger, Tom Henighan, Rewon Child, Aditya Ramesh, Daniel Ziegler, Jeffrey Wu, Clemens Winter, Chris Hesse, Mark Chen, Eric Sigler, Mateusz Litwin, Scott Gray, Benjamin Chess, Jack Clark, Christopher Berner, Sam McCandlish, Alec Radford, Ilya Sutskever, and Dario Amodei.
\newblock Language models are few-shot learners.
\newblock In H.~Larochelle, M.~Ranzato, R.~Hadsell, M.F. Balcan, and H.~Lin, editors, \emph{Advances in Neural Information Processing Systems}, volume~33, pages 1877--1901. Curran Associates, Inc., 2020.
\newblock URL \url{https://proceedings.neurips.cc/paper_files/paper/2020/file/1457c0d6bfcb4967418bfb8ac142f64a-Paper.pdf}.

\bibitem[Bubeck et~al.(2009)Bubeck, Munos, and Stoltz]{bubeck2009pure}
S{\'e}bastien Bubeck, R{\'e}mi Munos, and Gilles Stoltz.
\newblock Pure exploration in multi-armed bandits problems.
\newblock In \emph{International conference on Algorithmic learning theory}, pages 23--37. Springer, 2009.

\bibitem[Chowdhery et~al.(2023)Chowdhery, Narang, Devlin, Bosma, Mishra, Roberts, Barham, Chung, Sutton, Gehrmann, Schuh, Shi, Tsvyashchenko, Maynez, Rao, Barnes, Tay, Shazeer, Prabhakaran, Reif, Du, Hutchinson, Pope, Bradbury, Austin, Isard, Gur-Ari, Yin, Duke, Levskaya, Ghemawat, Dev, Michalewski, Garcia, Misra, Robinson, Fedus, Zhou, Ippolito, Luan, Lim, Zoph, Spiridonov, Sepassi, Dohan, Agrawal, Omernick, Dai, Pillai, Pellat, Lewkowycz, Moreira, Child, Polozov, Lee, Zhou, Wang, Saeta, Diaz, Firat, Catasta, Wei, Meier-Hellstern, Eck, Dean, Petrov, and Fiedel]{10.5555/3648699.3648939}
Aakanksha Chowdhery, Sharan Narang, Jacob Devlin, Maarten Bosma, Gaurav Mishra, Adam Roberts, Paul Barham, Hyung~Won Chung, Charles Sutton, Sebastian Gehrmann, Parker Schuh, Kensen Shi, Sashank Tsvyashchenko, Joshua Maynez, Abhishek Rao, Parker Barnes, Yi~Tay, Noam Shazeer, Vinodkumar Prabhakaran, Emily Reif, Nan Du, Ben Hutchinson, Reiner Pope, James Bradbury, Jacob Austin, Michael Isard, Guy Gur-Ari, Pengcheng Yin, Toju Duke, Anselm Levskaya, Sanjay Ghemawat, Sunipa Dev, Henryk Michalewski, Xavier Garcia, Vedant Misra, Kevin Robinson, Liam Fedus, Denny Zhou, Daphne Ippolito, David Luan, Hyeontaek Lim, Barret Zoph, Alexander Spiridonov, Ryan Sepassi, David Dohan, Shivani Agrawal, Mark Omernick, Andrew~M. Dai, Thanumalayan~Sankaranarayana Pillai, Marie Pellat, Aitor Lewkowycz, Erica Moreira, Rewon Child, Oleksandr Polozov, Katherine Lee, Zongwei Zhou, Xuezhi Wang, Brennan Saeta, Mark Diaz, Orhan Firat, Michele Catasta, Jason Wei, Kathy Meier-Hellstern, Douglas Eck, Jeff Dean, Slav Petrov, and Noah Fiedel.
\newblock Palm: scaling language modeling with pathways.
\newblock \emph{J. Mach. Learn. Res.}, 24\penalty0 (1), January 2023.
\newblock ISSN 1532-4435.

\bibitem[Cobbe et~al.(2021)Cobbe, Kosaraju, Bavarian, Chen, Jun, Kaiser, Plappert, Tworek, Hilton, Nakano, Hesse, and Schulman]{cobbe2021trainingverifierssolvemath}
Karl Cobbe, Vineet Kosaraju, Mohammad Bavarian, Mark Chen, Heewoo Jun, Lukasz Kaiser, Matthias Plappert, Jerry Tworek, Jacob Hilton, Reiichiro Nakano, Christopher Hesse, and John Schulman.
\newblock Training verifiers to solve math word problems, 2021.
\newblock URL \url{https://arxiv.org/abs/2110.14168}.

\bibitem[Damani et~al.(2025)Damani, Shenfeld, Peng, Bobu, and Andreas]{damani2024learning}
Mehul Damani, Idan Shenfeld, Andi Peng, Andreea Bobu, and Jacob Andreas.
\newblock Learning how hard to think: Input-adaptive allocation of {LM} computation.
\newblock In \emph{The Thirteenth International Conference on Learning Representations}, 2025.

\bibitem[DeepSeek-AI et~al.(2025)DeepSeek-AI, Guo, Yang, Zhang, Song, Zhang, Xu, Zhu, Ma, Wang, Bi, Zhang, Yu, Wu, Wu, Gou, Shao, Li, Gao, Liu, Xue, Wang, Wu, Feng, Lu, Zhao, Deng, Zhang, Ruan, Dai, Chen, Ji, Li, Lin, Dai, Luo, Hao, Chen, Li, Zhang, Bao, Xu, Wang, Ding, Xin, Gao, Qu, Li, Guo, Li, Wang, Chen, Yuan, Qiu, Li, Cai, Ni, Liang, Chen, Dong, Hu, Gao, Guan, Huang, Yu, Wang, Zhang, Zhao, Wang, Zhang, Xu, Xia, Zhang, Zhang, Tang, Li, Wang, Li, Tian, Huang, Zhang, Wang, Chen, Du, Ge, Zhang, Pan, Wang, Chen, Jin, Chen, Lu, Zhou, Chen, Ye, Wang, Yu, Zhou, Pan, Li, Zhou, Wu, Ye, Yun, Pei, Sun, Wang, Zeng, Zhao, Liu, Liang, Gao, Yu, Zhang, Xiao, An, Liu, Wang, Chen, Nie, Cheng, Liu, Xie, Liu, Yang, Li, Su, Lin, Li, Jin, Shen, Chen, Sun, Wang, Song, Zhou, Wang, Shan, Li, Wang, Wei, Zhang, Xu, Li, Zhao, Sun, Wang, Yu, Zhang, Shi, Xiong, He, Piao, Wang, Tan, Ma, Liu, Guo, Ou, Wang, Gong, Zou, He, Xiong, Luo, You, Liu, Zhou, Zhu, Xu, Huang, Li, Zheng, Zhu, Ma, Tang, Zha, Yan, Ren, Ren, Sha, Fu, Xu, Xie, Zhang,
  Hao, Ma, Yan, Wu, Gu, Zhu, Liu, Li, Xie, Song, Pan, Huang, Xu, Zhang, and Zhang]{deepseekai2025deepseekr1incentivizingreasoningcapability}
DeepSeek-AI, Daya Guo, Dejian Yang, Haowei Zhang, Junxiao Song, Ruoyu Zhang, Runxin Xu, Qihao Zhu, Shirong Ma, Peiyi Wang, Xiao Bi, Xiaokang Zhang, Xingkai Yu, Yu~Wu, Z.~F. Wu, Zhibin Gou, Zhihong Shao, Zhuoshu Li, Ziyi Gao, Aixin Liu, Bing Xue, Bingxuan Wang, Bochao Wu, Bei Feng, Chengda Lu, Chenggang Zhao, Chengqi Deng, Chenyu Zhang, Chong Ruan, Damai Dai, Deli Chen, Dongjie Ji, Erhang Li, Fangyun Lin, Fucong Dai, Fuli Luo, Guangbo Hao, Guanting Chen, Guowei Li, H.~Zhang, Han Bao, Hanwei Xu, Haocheng Wang, Honghui Ding, Huajian Xin, Huazuo Gao, Hui Qu, Hui Li, Jianzhong Guo, Jiashi Li, Jiawei Wang, Jingchang Chen, Jingyang Yuan, Junjie Qiu, Junlong Li, J.~L. Cai, Jiaqi Ni, Jian Liang, Jin Chen, Kai Dong, Kai Hu, Kaige Gao, Kang Guan, Kexin Huang, Kuai Yu, Lean Wang, Lecong Zhang, Liang Zhao, Litong Wang, Liyue Zhang, Lei Xu, Leyi Xia, Mingchuan Zhang, Minghua Zhang, Minghui Tang, Meng Li, Miaojun Wang, Mingming Li, Ning Tian, Panpan Huang, Peng Zhang, Qiancheng Wang, Qinyu Chen, Qiushi Du, Ruiqi Ge, Ruisong
  Zhang, Ruizhe Pan, Runji Wang, R.~J. Chen, R.~L. Jin, Ruyi Chen, Shanghao Lu, Shangyan Zhou, Shanhuang Chen, Shengfeng Ye, Shiyu Wang, Shuiping Yu, Shunfeng Zhou, Shuting Pan, S.~S. Li, Shuang Zhou, Shaoqing Wu, Shengfeng Ye, Tao Yun, Tian Pei, Tianyu Sun, T.~Wang, Wangding Zeng, Wanjia Zhao, Wen Liu, Wenfeng Liang, Wenjun Gao, Wenqin Yu, Wentao Zhang, W.~L. Xiao, Wei An, Xiaodong Liu, Xiaohan Wang, Xiaokang Chen, Xiaotao Nie, Xin Cheng, Xin Liu, Xin Xie, Xingchao Liu, Xinyu Yang, Xinyuan Li, Xuecheng Su, Xuheng Lin, X.~Q. Li, Xiangyue Jin, Xiaojin Shen, Xiaosha Chen, Xiaowen Sun, Xiaoxiang Wang, Xinnan Song, Xinyi Zhou, Xianzu Wang, Xinxia Shan, Y.~K. Li, Y.~Q. Wang, Y.~X. Wei, Yang Zhang, Yanhong Xu, Yao Li, Yao Zhao, Yaofeng Sun, Yaohui Wang, Yi~Yu, Yichao Zhang, Yifan Shi, Yiliang Xiong, Ying He, Yishi Piao, Yisong Wang, Yixuan Tan, Yiyang Ma, Yiyuan Liu, Yongqiang Guo, Yuan Ou, Yuduan Wang, Yue Gong, Yuheng Zou, Yujia He, Yunfan Xiong, Yuxiang Luo, Yuxiang You, Yuxuan Liu, Yuyang Zhou, Y.~X. Zhu,
  Yanhong Xu, Yanping Huang, Yaohui Li, Yi~Zheng, Yuchen Zhu, Yunxian Ma, Ying Tang, Yukun Zha, Yuting Yan, Z.~Z. Ren, Zehui Ren, Zhangli Sha, Zhe Fu, Zhean Xu, Zhenda Xie, Zhengyan Zhang, Zhewen Hao, Zhicheng Ma, Zhigang Yan, Zhiyu Wu, Zihui Gu, Zijia Zhu, Zijun Liu, Zilin Li, Ziwei Xie, Ziyang Song, Zizheng Pan, Zhen Huang, Zhipeng Xu, Zhongyu Zhang, and Zhen Zhang.
\newblock Deepseek-r1: Incentivizing reasoning capability in llms via reinforcement learning, 2025.
\newblock URL \url{https://arxiv.org/abs/2501.12948}.

\bibitem[Du et~al.(2025)Du, Tian, Ronanki, Rongali, Bodapati, Galstyan, Wells, Schwartz, Huerta, and Peng]{du-etal-2025-context}
Yufeng Du, Minyang Tian, Srikanth Ronanki, Subendhu Rongali, Sravan~Babu Bodapati, Aram Galstyan, Azton Wells, Roy Schwartz, Eliu~A Huerta, and Hao Peng.
\newblock Context length alone hurts {LLM} performance despite perfect retrieval.
\newblock In Christos Christodoulopoulos, Tanmoy Chakraborty, Carolyn Rose, and Violet Peng, editors, \emph{Findings of the Association for Computational Linguistics: EMNLP 2025}, pages 23281--23298, Suzhou, China, November 2025. Association for Computational Linguistics.
\newblock ISBN 979-8-89176-335-7.
\newblock \doi{10.18653/v1/2025.findings-emnlp.1264}.
\newblock URL \url{https://aclanthology.org/2025.findings-emnlp.1264/}.

\bibitem[Hoffmann et~al.(2022)Hoffmann, Borgeaud, Mensch, Buchatskaya, Cai, Rutherford, de~Las~Casas, Hendricks, Welbl, Clark, Hennigan, Noland, Millican, van~den Driessche, Damoc, Guy, Osindero, Simonyan, Elsen, Vinyals, Rae, and Sifre]{10.5555/3600270.3602446}
Jordan Hoffmann, Sebastian Borgeaud, Arthur Mensch, Elena Buchatskaya, Trevor Cai, Eliza Rutherford, Diego de~Las~Casas, Lisa~Anne Hendricks, Johannes Welbl, Aidan Clark, Tom Hennigan, Eric Noland, Katie Millican, George van~den Driessche, Bogdan Damoc, Aurelia Guy, Simon Osindero, Karen Simonyan, Erich Elsen, Oriol Vinyals, Jack~W. Rae, and Laurent Sifre.
\newblock Training compute-optimal large language models.
\newblock In \emph{Proceedings of the 36th International Conference on Neural Information Processing Systems}, NIPS '22, Red Hook, NY, USA, 2022. Curran Associates Inc.
\newblock ISBN 9781713871088.

\bibitem[Jain et~al.(2025)Jain, Han, Gu, Li, Yan, Zhang, Wang, Solar-Lezama, Sen, and Stoica]{jain2025livecodebench}
Naman Jain, King Han, Alex Gu, Wen-Ding Li, Fanjia Yan, Tianjun Zhang, Sida Wang, Armando Solar-Lezama, Koushik Sen, and Ion Stoica.
\newblock Livecodebench: Holistic and contamination free evaluation of large language models for code.
\newblock In \emph{The Thirteenth International Conference on Learning Representations}, 2025.
\newblock URL \url{https://openreview.net/forum?id=chfJJYC3iL}.

\bibitem[Jamieson and Nowak(2014)]{jamieson2014best}
Kevin Jamieson and Robert Nowak.
\newblock Best-arm identification algorithms for multi-armed bandits in the fixed confidence setting.
\newblock In \emph{2014 48th annual conference on information sciences and systems (CISS)}, pages 1--6. IEEE, 2014.

\bibitem[Kalai et~al.(2025)Kalai, Nachum, Vempala, and Zhang]{kalai2025languagemodelshallucinate}
Adam~Tauman Kalai, Ofir Nachum, Santosh~S. Vempala, and Edwin Zhang.
\newblock Why language models hallucinate, 2025.
\newblock URL \url{https://arxiv.org/abs/2509.04664}.

\bibitem[Kaplan et~al.(2020)Kaplan, McCandlish, Henighan, Brown, Chess, Child, Gray, Radford, Wu, and Amodei]{kaplan2020scalinglawsneurallanguage}
Jared Kaplan, Sam McCandlish, Tom Henighan, Tom~B. Brown, Benjamin Chess, Rewon Child, Scott Gray, Alec Radford, Jeffrey Wu, and Dario Amodei.
\newblock Scaling laws for neural language models, 2020.
\newblock URL \url{https://arxiv.org/abs/2001.08361}.

\bibitem[Kojima et~al.(2022)Kojima, Gu, Reid, Matsuo, and Iwasawa]{10.5555/3600270.3601883}
Takeshi Kojima, Shixiang~Shane Gu, Machel Reid, Yutaka Matsuo, and Yusuke Iwasawa.
\newblock Large language models are zero-shot reasoners.
\newblock In \emph{Proceedings of the 36th International Conference on Neural Information Processing Systems}, NIPS '22, Red Hook, NY, USA, 2022. Curran Associates Inc.
\newblock ISBN 9781713871088.

\bibitem[Lambert et~al.(2025)Lambert, Pyatkin, Morrison, Miranda, Lin, Chandu, Dziri, Kumar, Zick, Choi, Smith, and Hajishirzi]{lambert-etal-2025-rewardbench}
Nathan Lambert, Valentina Pyatkin, Jacob Morrison, LJ~Miranda, Bill~Yuchen Lin, Khyathi Chandu, Nouha Dziri, Sachin Kumar, Tom Zick, Yejin Choi, Noah~A. Smith, and Hannaneh Hajishirzi.
\newblock {R}eward{B}ench: Evaluating reward models for language modeling, April 2025.
\newblock URL \url{https://aclanthology.org/2025.findings-naacl.96/}.

\bibitem[Lewkowycz et~al.(2022)Lewkowycz, Andreassen, Dohan, Dyer, Michalewski, Ramasesh, Slone, Anil, Schlag, Gutman-Solo, Wu, Neyshabur, Gur-Ari, and Misra]{10.5555/3600270.3600548}
Aitor Lewkowycz, Anders Andreassen, David Dohan, Ethan Dyer, Henryk Michalewski, Vinay Ramasesh, Ambrose Slone, Cem Anil, Imanol Schlag, Theo Gutman-Solo, Yuhuai Wu, Behnam Neyshabur, Guy Gur-Ari, and Vedant Misra.
\newblock Solving quantitative reasoning problems with language models.
\newblock In \emph{Proceedings of the 36th International Conference on Neural Information Processing Systems}, NIPS '22, Red Hook, NY, USA, 2022. Curran Associates Inc.
\newblock ISBN 9781713871088.

\bibitem[Lightman et~al.(2024)Lightman, Kosaraju, Burda, Edwards, Baker, Lee, Leike, Schulman, Sutskever, and Cobbe]{lightman2024lets}
Hunter Lightman, Vineet Kosaraju, Yuri Burda, Harrison Edwards, Bowen Baker, Teddy Lee, Jan Leike, John Schulman, Ilya Sutskever, and Karl Cobbe.
\newblock Let's verify step by step.
\newblock In \emph{The Twelfth International Conference on Learning Representations}, 2024.
\newblock URL \url{https://openreview.net/forum?id=v8L0pN6EOi}.

\bibitem[Lin et~al.(2025)Lin, Xu, Hu, Li, Hao, Zhang, and Cai]{10.1007/s10489-025-07044-6}
Qingwen Lin, Boyan Xu, Guimin Hu, Zijian Li, Zhifeng Hao, Keli Zhang, and Ruichu Cai.
\newblock Cmcts: A constrained monte carlo tree search framework for mathematical reasoning in large language model, December 2025.
\newblock ISSN 0924-669X.
\newblock URL \url{https://doi.org/10.1007/s10489-025-07044-6}.

\bibitem[Liu et~al.(2021)Liu, Shen, Zhang, Dolan, Carin, and Chen]{liu2021makesgoodincontextexamples}
Jiachang Liu, Dinghan Shen, Yizhe Zhang, Bill Dolan, Lawrence Carin, and Weizhu Chen.
\newblock What makes good in-context examples for gpt-$3$?, 2021.
\newblock URL \url{https://arxiv.org/abs/2101.06804}.

\bibitem[Liu et~al.(2025)Liu, Gao, Zhao, Zhang, Li, Qi, Ouyang, and Zhou]{liu2025can}
Runze Liu, Junqi Gao, Jian Zhao, Kaiyan Zhang, Xiu Li, Biqing Qi, Wanli Ouyang, and Bowen Zhou.
\newblock Can 1b {LLM} surpass 405b {LLM}? rethinking compute-optimal test-time scaling.
\newblock In \emph{Workshop on Reasoning and Planning for Large Language Models}, 2025.
\newblock URL \url{https://openreview.net/forum?id=CvjX9Lhpze}.

\bibitem[Locatelli et~al.(2016)Locatelli, Gutzeit, and Carpentier]{locatelli2016optimalalgorithmthresholdingbandit}
Andrea Locatelli, Maurilio Gutzeit, and Alexandra Carpentier.
\newblock An optimal algorithm for the thresholding bandit problem, 2016.
\newblock URL \url{https://arxiv.org/abs/1605.08671}.

\bibitem[Madaan et~al.(2023)Madaan, Tandon, Gupta, Hallinan, Gao, Wiegreffe, Alon, Dziri, Prabhumoye, Yang, Gupta, Majumder, Hermann, Welleck, Yazdanbakhsh, and Clark]{madaan2023selfrefine}
Aman Madaan, Niket Tandon, Prakhar Gupta, Skyler Hallinan, Luyu Gao, Sarah Wiegreffe, Uri Alon, Nouha Dziri, Shrimai Prabhumoye, Yiming Yang, Shashank Gupta, Bodhisattwa~Prasad Majumder, Katherine Hermann, Sean Welleck, Amir Yazdanbakhsh, and Peter Clark.
\newblock Self-refine: Iterative refinement with self-feedback.
\newblock In \emph{Thirty-seventh Conference on Neural Information Processing Systems}, 2023.
\newblock URL \url{https://openreview.net/forum?id=S37hOerQLB}.

\bibitem[Manvi et~al.(2024)Manvi, Singh, and Ermon]{manvi2024adaptiveinferencetimecomputellms}
Rohin Manvi, Anikait Singh, and Stefano Ermon.
\newblock Adaptive inference-time compute: Llms can predict if they can do better, even mid-generation, 2024.
\newblock URL \url{https://arxiv.org/abs/2410.02725}.

\bibitem[Mosbach et~al.(2023)Mosbach, Pimentel, Ravfogel, Klakow, and Elazar]{mosbach-etal-2023-shot}
Marius Mosbach, Tiago Pimentel, Shauli Ravfogel, Dietrich Klakow, and Yanai Elazar.
\newblock Few-shot fine-tuning vs. in-context learning: A fair comparison and evaluation.
\newblock In Anna Rogers, Jordan Boyd-Graber, and Naoaki Okazaki, editors, \emph{Findings of the Association for Computational Linguistics: ACL 2023}, pages 12284--12314, Toronto, Canada, July 2023. Association for Computational Linguistics.
\newblock \doi{10.18653/v1/2023.findings-acl.779}.
\newblock URL \url{https://aclanthology.org/2023.findings-acl.779/}.

\bibitem[Muennighoff et~al.(2025)Muennighoff, Yang, Shi, Li, Fei-Fei, Hajishirzi, Zettlemoyer, Liang, Cand{\`e}s, and Hashimoto]{muennighoff2025s1}
Niklas Muennighoff, Zitong Yang, Weijia Shi, Xiang~Lisa Li, Li~Fei-Fei, Hannaneh Hajishirzi, Luke Zettlemoyer, Percy Liang, Emmanuel Cand{\`e}s, and Tatsunori~B Hashimoto.
\newblock s1: Simple test-time scaling.
\newblock In \emph{Proceedings of the 2025 Conference on Empirical Methods in Natural Language Processing}, pages 20286--20332, 2025.

\bibitem[OpenAI(2024)]{openai2024learning}
OpenAI.
\newblock Learning to reason with llms, 2024.
\newblock URL \url{https://openai.com/index/learning-to-reason-with-llms/}.
\newblock Accessed 12-Sep-2024.

\bibitem[Qin et~al.(2024)Qin, Zhang, Chen, Dagar, and Ye]{qin2024context}
Chengwei Qin, Aston Zhang, Chen Chen, Anirudh Dagar, and Wenming Ye.
\newblock In-context learning with iterative demonstration selection.
\newblock In \emph{Findings of the Association for Computational Linguistics: EMNLP 2024}, pages 7441--7455, 2024.

\bibitem[Rein et~al.(2024)Rein, Hou, Stickland, Petty, Pang, Dirani, Michael, and Bowman]{rein2024gpqa}
David Rein, Betty~Li Hou, Asa~Cooper Stickland, Jackson Petty, Richard~Yuanzhe Pang, Julien Dirani, Julian Michael, and Samuel~R Bowman.
\newblock Gpqa: A graduate-level google-proof q\&a benchmark.
\newblock In \emph{First conference on language modeling}, 2024.

\bibitem[Snell et~al.(2025)Snell, Lee, Xu, and Kumar]{snell2025scaling}
Charlie~Victor Snell, Jaehoon Lee, Kelvin Xu, and Aviral Kumar.
\newblock Scaling {LLM} test-time compute optimally can be more effective than scaling parameters for reasoning.
\newblock In \emph{The Thirteenth International Conference on Learning Representations}, 2025.
\newblock URL \url{https://openreview.net/forum?id=4FWAwZtd2n}.

\bibitem[Stojanovski et~al.(2025)Stojanovski, Stanley, Sharratt, Jones, Adefioye, Kaddour, and K{\"o}pf]{stojanovski2025reasoning}
Zafir Stojanovski, Oliver Stanley, Joe Sharratt, Richard Jones, Abdulhakeem Adefioye, Jean Kaddour, and Andreas K{\"o}pf.
\newblock Reasoning gym: Reasoning environments for reinforcement learning with verifiable rewards.
\newblock In \emph{The Thirty-ninth Annual Conference on Neural Information Processing Systems Datasets and Benchmarks Track}, 2025.
\newblock URL \url{https://openreview.net/forum?id=GqYSunGmp7}.

\bibitem[Sun et~al.(2024)Sun, Haider, Zhang, Yang, Qiu, Yin, Wang, Bartlett, and Zanette]{sun2024fast}
Hanshi Sun, Momin Haider, Ruiqi Zhang, Huitao Yang, Jiahao Qiu, Ming Yin, Mengdi Wang, Peter Bartlett, and Andrea Zanette.
\newblock Fast best-of-n decoding via speculative rejection.
\newblock In \emph{The Thirty-eighth Annual Conference on Neural Information Processing Systems}, 2024.
\newblock URL \url{https://openreview.net/forum?id=348hfcprUs}.

\bibitem[Tan et~al.(2025)Tan, Zhang, Hu, Pan, and Wang]{tan2025adaptiverectificationsamplingtesttime}
Zhendong Tan, Xingjun Zhang, Chaoyi Hu, Yancheng Pan, and Shaoxun Wang.
\newblock Adaptive rectification sampling for test-time compute scaling, 2025.
\newblock URL \url{https://arxiv.org/abs/2504.01317}.

\bibitem[Tanwar et~al.(2023)Tanwar, Dutta, Borthakur, and Chakraborty]{tanwar-etal-2023-multilingual}
Eshaan Tanwar, Subhabrata Dutta, Manish Borthakur, and Tanmoy Chakraborty.
\newblock Multilingual {LLM}s are better cross-lingual in-context learners with alignment.
\newblock In Anna Rogers, Jordan Boyd-Graber, and Naoaki Okazaki, editors, \emph{Proceedings of the 61st Annual Meeting of the Association for Computational Linguistics (Volume 1: Long Papers)}, pages 6292--6307, Toronto, Canada, July 2023. Association for Computational Linguistics.
\newblock \doi{10.18653/v1/2023.acl-long.346}.
\newblock URL \url{https://aclanthology.org/2023.acl-long.346/}.

\bibitem[Team(2025)]{comanici2025gemini25pushingfrontier}
Google~Gemini Team.
\newblock Gemini 2.5: Pushing the frontier with advanced reasoning, multimodality, long context, and next generation agentic capabilities, 2025.
\newblock URL \url{https://arxiv.org/abs/2507.06261}.

\bibitem[Tuyls et~al.(2026)Tuyls, Foster, Krishnamurthy, and Ash]{tuyls2026representationbased}
Jens Tuyls, Dylan~J Foster, Akshay Krishnamurthy, and Jordan~T. Ash.
\newblock Representation-based exploration for language models: From test-time to post-training.
\newblock In \emph{The Fourteenth International Conference on Learning Representations}, 2026.
\newblock URL \url{https://openreview.net/forum?id=S4PCF1YxoR}.

\bibitem[Uesato et~al.(2022)Uesato, Kushman, Kumar, Song, Siegel, Wang, Creswell, Irving, and Higgins]{uesato2022solving}
Jonathan Uesato, Nate Kushman, Ramana Kumar, Francis Song, Noah Siegel, Lisa Wang, Antonia Creswell, Geoffrey Irving, and Irina Higgins.
\newblock Solving math word problems with process-and outcome-based feedback.
\newblock \emph{arXiv preprint arXiv:2211.14275}, 2022.

\bibitem[Wang et~al.(2025{\natexlab{a}})Wang, Zhu, Saad-Falcon, Athiwaratkun, Wu, Wang, Song, Zhang, Dhingra, and Zou]{wang2025thinkdeepthinkfast}
Junlin Wang, Shang Zhu, Jon Saad-Falcon, Ben Athiwaratkun, Qingyang Wu, Jue Wang, Shuaiwen~Leon Song, Ce~Zhang, Bhuwan Dhingra, and James Zou.
\newblock Think deep, think fast: Investigating efficiency of verifier-free inference-time-scaling methods, 2025{\natexlab{a}}.
\newblock URL \url{https://arxiv.org/abs/2504.14047}.

\bibitem[Wang et~al.(2025{\natexlab{b}})Wang, Feng, Li, Yuan, Zhang, Tan, Pan, Hu, and Li]{wang-etal-2025-make}
Xinglin Wang, Shaoxiong Feng, Yiwei Li, Peiwen Yuan, Yueqi Zhang, Chuyi Tan, Boyuan Pan, Yao Hu, and Kan Li.
\newblock Make every penny count: Difficulty-adaptive self-consistency for cost-efficient reasoning.
\newblock In Luis Chiruzzo, Alan Ritter, and Lu~Wang, editors, \emph{Findings of the Association for Computational Linguistics: NAACL 2025}, pages 6919--6932, Albuquerque, New Mexico, April 2025{\natexlab{b}}. Association for Computational Linguistics.
\newblock ISBN 979-8-89176-195-7.
\newblock \doi{10.18653/v1/2025.findings-naacl.383}.
\newblock URL \url{https://aclanthology.org/2025.findings-naacl.383/}.

\bibitem[Wang et~al.(2023)Wang, Wei, Schuurmans, Le, Chi, Narang, Chowdhery, and Zhou]{wang2023selfconsistency}
Xuezhi Wang, Jason Wei, Dale Schuurmans, Quoc~V Le, Ed~H. Chi, Sharan Narang, Aakanksha Chowdhery, and Denny Zhou.
\newblock Self-consistency improves chain of thought reasoning in language models.
\newblock In \emph{The Eleventh International Conference on Learning Representations}, 2023.
\newblock URL \url{https://openreview.net/forum?id=1PL1NIMMrw}.

\bibitem[Wei et~al.(2022)Wei, Wang, Schuurmans, Bosma, brian ichter, Xia, Chi, Le, and Zhou]{wei2022chain}
Jason Wei, Xuezhi Wang, Dale Schuurmans, Maarten Bosma, brian ichter, Fei Xia, Ed~H. Chi, Quoc~V Le, and Denny Zhou.
\newblock Chain of thought prompting elicits reasoning in large language models.
\newblock In Alice~H. Oh, Alekh Agarwal, Danielle Belgrave, and Kyunghyun Cho, editors, \emph{Advances in Neural Information Processing Systems}, 2022.
\newblock URL \url{https://openreview.net/forum?id=_VjQlMeSB_J}.

\bibitem[Wu et~al.(2025)Wu, Mirhoseini, and Tambe]{wu2025roletemperaturesamplingtesttime}
Yuheng Wu, Azalia Mirhoseini, and Thierry Tambe.
\newblock On the role of temperature sampling in test-time scaling, 2025.
\newblock URL \url{https://arxiv.org/abs/2510.02611}.

\bibitem[Xia et~al.(2025)Xia, Luo, Bartels, Xu, and Li]{xia2025rethinkingunsolvableincontextsearch}
Fanzeng Xia, Yidong Luo, Tinko~Sebastian Bartels, Yaqi Xu, and Tongxin Li.
\newblock Rethinking the unsolvable: When in-context search meets test-time scaling, 2025.
\newblock URL \url{https://arxiv.org/abs/2505.22290}.

\bibitem[Xu et~al.(2023)Xu, Wang, Mao, Lyu, She, and Zhang]{xu2023knn}
Benfeng Xu, Quan Wang, Zhendong Mao, Yajuan Lyu, Qiaoqiao She, and Yongdong Zhang.
\newblock \$k\${NN} prompting: Beyond-context learning with calibration-free nearest neighbor inference.
\newblock In \emph{The Eleventh International Conference on Learning Representations}, 2023.
\newblock URL \url{https://openreview.net/forum?id=fe2S7736sNS}.

\bibitem[Xu et~al.(2025)Xu, Jain, and Kankanhalli]{xu2025hallucinationinevitableinnatelimitation}
Ziwei Xu, Sanjay Jain, and Mohan Kankanhalli.
\newblock Hallucination is inevitable: An innate limitation of large language models, 2025.
\newblock URL \url{https://arxiv.org/abs/2401.11817}.

\bibitem[Yao et~al.(2023)Yao, Yu, Zhao, Shafran, Griffiths, Cao, and Narasimhan]{yao2023tree}
Shunyu Yao, Dian Yu, Jeffrey Zhao, Izhak Shafran, Thomas~L. Griffiths, Yuan Cao, and Karthik~R Narasimhan.
\newblock Tree of thoughts: Deliberate problem solving with large language models.
\newblock In \emph{Thirty-seventh Conference on Neural Information Processing Systems}, 2023.
\newblock URL \url{https://openreview.net/forum?id=5Xc1ecxO1h}.

\bibitem[Yoran et~al.(2024)Yoran, Wolfson, Ram, and Berant]{yoran2024making}
Ori Yoran, Tomer Wolfson, Ori Ram, and Jonathan Berant.
\newblock Making retrieval-augmented language models robust to irrelevant context.
\newblock In \emph{The Twelfth International Conference on Learning Representations}, 2024.
\newblock URL \url{https://openreview.net/forum?id=ZS4m74kZpH}.

\bibitem[Zhang et~al.(2025)Zhang, Zheng, Wu, Zhang, Lin, Yu, Liu, Zhou, and Lin]{zhang-etal-2025-lessons}
Zhenru Zhang, Chujie Zheng, Yangzhen Wu, Beichen Zhang, Runji Lin, Bowen Yu, Dayiheng Liu, Jingren Zhou, and Junyang Lin.
\newblock The lessons of developing process reward models in mathematical reasoning.
\newblock In Wanxiang Che, Joyce Nabende, Ekaterina Shutova, and Mohammad~Taher Pilehvar, editors, \emph{Findings of the Association for Computational Linguistics: ACL 2025}, pages 10495--10516, Vienna, Austria, July 2025. Association for Computational Linguistics.
\newblock ISBN 979-8-89176-256-5.
\newblock \doi{10.18653/v1/2025.findings-acl.547}.
\newblock URL \url{https://aclanthology.org/2025.findings-acl.547/}.

\bibitem[Zhu et~al.(2020)Zhu, Katariya, and Nowak]{zhu2020robust}
Yinglun Zhu, Sumeet Katariya, and Robert Nowak.
\newblock Robust outlier arm identification.
\newblock In \emph{International Conference on Machine Learning}, pages 11566--11575. PMLR, 2020.

\bibitem[Zhu et~al.(2021)Zhu, Zhou, Jiang, Gu, Willett, and Nowak]{zhu2021pure}
Yinglun Zhu, Dongruo Zhou, Ruoxi Jiang, Quanquan Gu, Rebecca Willett, and Robert Nowak.
\newblock Pure exploration in kernel and neural bandits.
\newblock \emph{Advances in neural information processing systems}, 34:\penalty0 11618--11630, 2021.

\bibitem[Zhu et~al.(2022)Zhu, Katz-Samuels, and Nowak]{zhu2022near}
Yinglun Zhu, Julian Katz-Samuels, and Robert Nowak.
\newblock Near instance optimal model selection for pure exploration linear bandits.
\newblock In \emph{International Conference on Artificial Intelligence and Statistics}, pages 6735--6769. PMLR, 2022.

\bibitem[Zuo and Zhu(2026)]{zuo2025strategicscalingtesttimecompute}
Bowen Zuo and Yinglun Zhu.
\newblock Strategic scaling of test-time compute: A bandit learning approach.
\newblock In \emph{The Fourteenth International Conference on Learning Representations}, 2026.

\end{thebibliography}

\newpage
\appendix
\section{Other details for experiments}
\label{app:experiments}
\paragraph{Reasoning-Gym Dataset Construction.}
To evaluate our method on diverse forms of symbolic and algorithmic reasoning, we construct a balanced benchmark using the Reasoning Gym \citep{stojanovski2025reasoning} framework. We focus on four representative categories that capture complementary reasoning skills: \emph{decimal arithmetic}, \emph{advanced geometry}, \emph{binary alternation}, and \emph{letter counting}. For each category, we sample 100 problem instances using a fixed random seed to ensure reproducibility, resulting in a total of 400 evaluation problems.

For each category, we instantiate the corresponding Reasoning Gym task using its canonical dataset interface and verify correctness using the task-specific scoring function provided by the framework. All generated instances include the problem statement, ground-truth answer, and metadata identifying the originating task. We aggregate the resulting problems into a unified evaluation set, which is used consistently across all methods and experimental settings.

\paragraph{Model APIs and decoding settings.}
All experiments are conducted using publicly available LLM APIs.
Unless otherwise specified, response generation uses a temperature of $0.3$ when temperature control is supported.
We set $\texttt{Top\_P}=0.9$ and $\texttt{Top\_K}=40$ for all models that expose these parameters.

For \texttt{Gemini} family models, we use the maximum available output token limit of $65{,}536$ tokens.
When \emph{thinking} is enabled, we allocate a maximum thinking budget of $24{,}576$ tokens.
For all \texttt{GPT} family models, we set the maximum output length to $32{,}768$ tokens. The reasoning and verbosity levels are all medium. To get the question embeddings, we mainly use Gemini's text embedding model.  

\paragraph{Prompt construction for MATH-500.}
We used the same prompt across different models.  
 Our method uses two prompt variants: a \emph{warm-up prompt} used for initial sampling, and an \emph{adaptive prompt} that prepends $P$ semantically similar test-time demonstrations (neighbors) before querying the target problem. Illustrative examples used in our MATH-500 experiment are shown below.

The warm-up stage uses a zero-shot CoT \citep{10.5555/3600270.3601883} prompt shared across all models.

\begin{tcolorbox}[
    colback=blue!5!white,
    colframe=blue!75!black,
    width=\linewidth,
    arc=2mm,
    boxrule=1pt,
    left=3mm, right=3mm,
    top=2mm, bottom=2mm,
    title=\textbf{MATH-500 warm-up prompt}
]
\noindent
Reason step by step and put the final answer in \textbackslash boxed\{\}.\\
Problem:\\
\{\textit{Problem statement}\}

\medskip
Solution:
\end{tcolorbox}

We do not apply any complicated prompting engineering at this stage. We simply feed in the previously generated $D_{test}$ demonstrations into the prompt constructions.
Each example consists of a previously generated question–solution pair.
The target question is appended after a separator.

\begin{tcolorbox}[
    colback=orange!5!white,
    colframe=orange!80!black,
    width=\linewidth,
    arc=2mm,
    boxrule=1pt,
    left=3mm, right=3mm,
    top=2mm, bottom=2mm,
    title=\textbf{MATH-500 adaptive prompt}
]
\noindent
Example 1\\
Problem:\\
\{\textit{Neighbor problem 1}\}

\medskip
Solution:\\
\{\textit{Neighbor solution 1}\}

\medskip
Example 2\\
Problem:\\
\{\textit{Neighbor problem 2}\}

\medskip
Solution:\\
\{\textit{Neighbor solution 2}\}

\medskip
Example 3\\
Problem:\\
\{\textit{Neighbor problem 3}\}

\medskip
Solution:\\
\{\textit{Neighbor solution 3}\}

\medskip
\noindent\rule{\linewidth}{0.4pt}

\medskip
Now solve the following. Put the final answer in \textbackslash boxed\{\}.\\
Problem:\\
\{\textit{Target question}\}

\medskip
Solution:
\end{tcolorbox}

\paragraph{Prompt construction for LiveCodeBench.}
For LiveCodeBench, we use a code-only generation protocol: the model must output \emph{only} valid Python 3 source code (no explanations, no comments, and no markdown fences). We employ two prompt variants: (i) a \emph{base prompt} used in warm-up sampling and baseline settings, and (ii) an \emph{adaptive neighbor+base prompt} that prepends up to three previously solved test-time examples before appending the base prompt verbatim.

The base prompt provides the problem statement, optional starter code, and explicit I/O and environment constraints.

\begin{tcolorbox}[
    colback=blue!5!white,
    colframe=blue!75!black,
    width=\linewidth,
    arc=2mm,
    boxrule=1pt,
    left=3mm, right=3mm,
    top=2mm, bottom=2mm,
    title=\textbf{LiveCodeBench warm-up prompt}
]
\noindent
You are a competitive programming assistant.\\
Return ONLY valid Python 3 code that solves the task. Do not include explanations, comments, or markdown fences.

\medskip
Problem:\\
\{\textit{question\_content}\}

\medskip
Starter (optional):\\
\{\textit{starter\_code or N/A}\}

\medskip
I/O requirements:
\begin{itemize}
  \item Read from STDIN exactly as described in the problem.
  \item Write only the required answer(s) to STDOUT (no extra prints).
  \item Do not read/write files. Do not use network.
  \item Avoid heavy/obscure libraries.
\end{itemize}

Constraints:
\begin{itemize}
  \item Python 3.10+.
  \item Prefer iterative solutions if deep recursion could occur.
  \item If multiple test cases exist, handle them all.
\end{itemize}

\medskip
Return ONLY the Python source code.
\end{tcolorbox}

In the adaptive stage, we prepend previously solved coding problems and their Python solutions (selected from test-time generations), followed by the base prompt verbatim. This construction provides retrieval-style guidance while preserving the original task specification.

\begin{tcolorbox}[
    colback=orange!5!white,
    colframe=orange!80!black,
    width=\linewidth,
    arc=2mm,
    boxrule=1pt,
    left=3mm, right=3mm,
    top=2mm, bottom=2mm,
    title=\textbf{LiveCodeBench adaptive prompt}
]
\noindent
You will see up to 3 previously solved coding problems with their Python 3 solutions.\\
Use them as guidance. After the examples, follow the final instructions below.\\
Return ONLY valid Python 3 code (no explanations, comments, or markdown fences).

\medskip
\noindent\rule{\linewidth}{0.4pt}

\medskip
Example 1\\
Problem:\\
\{\textit{neighbor\_problem\_1}\}

\medskip
Solution (Python 3 code only):\\
\{\textit{neighbor\_solution\_1}\}

\medskip
\noindent\rule{\linewidth}{0.4pt}

\medskip
Example 2\\
Problem:\\
\{\textit{neighbor\_problem\_2}\}

\medskip
Solution (Python 3 code only):\\
\{\textit{neighbor\_solution\_2}\}

\medskip
\noindent\rule{\linewidth}{0.4pt}

\medskip
Example 3\\
Problem:\\
\{\textit{neighbor\_problem\_3}\}

\medskip
Solution (Python 3 code only):\\
\{\textit{neighbor\_solution\_3}\}

\medskip
\noindent\rule{\linewidth}{0.4pt}

\medskip
\textbf{Base prompt (appended verbatim):}\\
\{\textit{LiveCodeBench Base Prompt content}\}
\end{tcolorbox}

\paragraph{Prompt construction for MinervaMath.}
For MinervaMath, the setting is similar to that of the MATH-500's. 

\begin{tcolorbox}[
    colback=blue!5!white,
    colframe=blue!75!black,
    width=\linewidth,
    arc=2mm,
    boxrule=1pt,
    left=3mm, right=3mm,
    top=2mm, bottom=2mm,
    title=\textbf{MinervaMath warm-up prompt}
]
\noindent
Let's solve this step by step and put the final answer in \textbackslash boxed\{\} with just the numeric value.\\
Problem:\\
\{\textit{Problem statement}\}

\medskip
Solution:
\end{tcolorbox}

\begin{tcolorbox}[
    colback=orange!5!white,
    colframe=orange!80!black,
    width=\linewidth,
    arc=2mm,
    boxrule=1pt,
    left=3mm, right=3mm,
    top=2mm, bottom=2mm,
    title=\textbf{MinervaMath adaptive prompt}
]
\noindent
Example 1\\
Problem:\\
\{\textit{Neighbor problem 1}\}

\medskip
Solution:\\
\{\textit{Neighbor solution 1}\}

\medskip
Example 2\\
Problem:\\
\{\textit{Neighbor problem 2}\}

\medskip
Solution:\\
\{\textit{Neighbor solution 2}\}

\medskip
Example 3\\
Problem:\\
\{\textit{Neighbor problem 3}\}

\medskip
Solution:\\
\{\textit{Neighbor solution 3}\}

\medskip
\noindent\texttt{---}

\medskip
Let's solve this step by step and put the final answer in \textbackslash boxed\{\} with just the numeric value.\\
Problem:\\
\{\textit{Target problem}\}

\medskip
Solution:
\end{tcolorbox}
 
\paragraph{Prompt construction for GPQA-Diamond.}
For GPQA, we use a multiple-choice solving protocol that asks the model to reason step by step and output exactly one option letter in $\text{boxed\{\}}$ (A/B/C/D).

\begin{tcolorbox}[
    colback=blue!5!white,
    colframe=blue!75!black,
    width=\linewidth,
    arc=2mm,
    boxrule=1pt,
    left=3mm, right=3mm,
    top=2mm, bottom=2mm,
    title=\textbf{GPQA warm-up MCQ prompt}
]
\noindent
Solve the problem step by step.\\
Select one selection and answer A, B, C or D in \textbackslash boxed\{\}.

\medskip
Question: \{\textit{question\_text}\}
\end{tcolorbox}

\begin{tcolorbox}[
    colback=orange!5!white,
    colframe=orange!80!black,
    width=\linewidth,
    arc=2mm,
    boxrule=1pt,
    left=3mm, right=3mm,
    top=2mm, bottom=2mm,
    title=\textbf{GPQA adaptive prompt}
]
\noindent
You will see up to 3 previously answered multiple-choice questions.\\
Use them as guidance. After the examples, solve the question step by step.\\
Select one selection and answer A, B, C or D in \textbackslash boxed\{\}.

\medskip
\noindent\texttt{---}

\medskip
Example 1\\
\{\textit{neighbor\_question\_1}\}\\
Answer: \{\textit{neighbor\_answer\_1}\}

\medskip
\noindent\texttt{---}

\medskip
Example 2\\
\{\textit{neighbor\_question\_2}\}\\
Answer: \{\textit{neighbor\_answer\_2}\}

\medskip
\noindent\texttt{---}

\medskip
Example 3\\
\{\textit{neighbor\_question\_3}\}\\
Answer: \{\textit{neighbor\_answer\_3}\}

\medskip
\noindent\texttt{---}

\medskip
Solve the problem step by step.\\
Select one selection and answer A, B, C or D in \textbackslash boxed\{\}.

\medskip
Question: \{\textit{target\_question\_text}\}
\end{tcolorbox}

\paragraph{Prompt construction for Reasoning Gym.}
For Reasoning Gym, we use a minimal answer-format instruction that requires the final answer to be enclosed in $\text{boxed\{\}}$.
In contrast to math benchmarks, where explicit zero-shot CoT prompting is commonly used, we find it not useful in this setting. Thus, we apply a prompt without it.

\begin{tcolorbox}[
    colback=blue!5!white,
    colframe=blue!75!black,
    width=\linewidth,
    arc=2mm,
    boxrule=1pt,
    left=3mm, right=3mm,
    top=2mm, bottom=2mm,
    title=\textbf{Reasoning Gym warm-up prompt}
]
\noindent
Put the final answer within \textbackslash boxed\{\}.\\
Problem:\\
\{\textit{Problem statement}\}

\medskip
Solution:
\end{tcolorbox}

\begin{tcolorbox}[
    colback=orange!5!white,
    colframe=orange!80!black,
    width=\linewidth,
    arc=2mm,
    boxrule=1pt,
    left=3mm, right=3mm,
    top=2mm, bottom=2mm,
    title=\textbf{Reasoning Gym adaptive prompt}
]
\noindent
Example 1\\
Problem:\\
\{\textit{Neighbor problem 1}\}

\medskip
Solution:\\
\{\textit{Neighbor solution 1}\}

\medskip
Example 2\\
Problem:\\
\{\textit{Neighbor problem 2}\}

\medskip
Solution:\\
\{\textit{Neighbor solution 2}\}

\medskip
Example 3\\
Problem:\\
\{\textit{Neighbor problem 3}\}

\medskip
Solution:\\
\{\textit{Neighbor solution 3}\}

\medskip
\noindent\texttt{---}

\medskip
Now solve the following. Put the final answer within \textbackslash boxed\{\}.\\
Problem:\\
\{\textit{Target question}\}

\medskip
Solution:
\end{tcolorbox}

\section{Additional experimental results}
\label{app:additional_experiments}
\paragraph{Additional allocation method.}
We incorporate an additional adaptive baseline inspired by \citep{wang-etal-2025-make}, referred to as Difficulty-Aware allocation \textsc{DA}. In this implementation, we use Gemini-3-flash to estimate per-question difficulty and allocate computation proportionally. Unlike the original batch-based formulation in \citep{wang-etal-2025-make}, we perform per-query difficulty estimation, which allows a more fine-grained allocation strategy aligned with our test-time adaptive setting.

To ensure a fair comparison, we strictly control the computational budget. Specifically, we maintain approximately the same total token usage as the Elim baseline, so that any performance difference reflects allocation strategy rather than increased computation. Given that token usage is matched and fully reported, we present results primarily across rounds to highlight differences in allocation dynamics. We can see that our approach still outperforms all other variations.
\begin{table}[t]
\caption{Coverage (\%) across rounds on MinervaMath using Gemini-2.5-flash-lite (thinking).}
\centering
\small
\setlength{\tabcolsep}{6pt}
\renewcommand{\arraystretch}{1.1}
\begin{tabular}{lcccc}
\toprule
Method & R0 & R1 & R2 & R3 \\
\midrule
DA          & 41.5 & 48.2 & 51.1 & 51.5 \\
Elim        & 41.5 & 47.8 & 51.8 & 52.9 \\
Ours (Rand) & 41.5 & 48.9 & 50.7 & 52.2 \\
Ours (Adpt) & 41.5 & 60.3 & 64.3 & 66.2 \\
\bottomrule
\end{tabular}
\label{tab:minerva_rounds}
\end{table}

\begin{table}[t]
\caption{Coverage (\%) across rounds on MATH500 using GPT-4.1-mini.}
\centering
\small
\setlength{\tabcolsep}{6pt}
\renewcommand{\arraystretch}{1.1}
\begin{tabular}{lcccc}
\toprule
Method & R0 & R1 & R2 & R3 \\
\midrule
DA          & 88.8 & 90.6 & 91.8 & 92.2 \\
Elim        & 88.8 & 91.0 & 92.4 & 92.4 \\
Ours (Rand) & 88.8 & 91.2 & 93.2 & 94.4 \\
Ours (Adpt) & 88.8 & 92.8 & 93.8 & 95.6 \\
\bottomrule
\end{tabular}
\label{tab:math500_rounds}
\end{table}

\paragraph{Token count for all usage.}
We select GPQA-Diamond, using \texttt{Gemini-2.5-flash-lite} model with thinking enabled, as a representative benchmark for token-usage analysis, since it has the longest average output token length among all evaluated datasets. The setting is identical to our main experiments. The results, reported in \cref{tab:tokens_round_avg}, show that our method significantly reduces token consumption at every round. Importantly, because our approach also yields higher coverage, the true efficiency gains are underestimated by token counts alone.

\begin{table}[t]
\caption{Average token usage (prompt + thinking + response) per round, across 4 seeds.}
\centering
\small
\setlength{\tabcolsep}{6pt}
\renewcommand{\arraystretch}{1.1}
\begin{tabular}{lrrrrr}
\toprule
Method & R0 & R1 & R2 & R3 & Total \\
\midrule
Elim  & 2,025,345 & 1,680,173 & 1,308,144 & 1,249,419 & 6,263,081 \\
Ours  & 2,025,345 & 1,102,486 &   902,109 &   741,611 & 4,771,551 \\
\midrule
Savings & 0 & 577,687 & 406,035 & 507,808 & 1,491,530 \\
\bottomrule
\end{tabular}
\label{tab:tokens_round_avg}
\end{table}

\begin{figure*}[ht]
  \centering
  \begin{subfigure}[t]{0.48\linewidth}
    \centering
    \includegraphics[width=\linewidth]{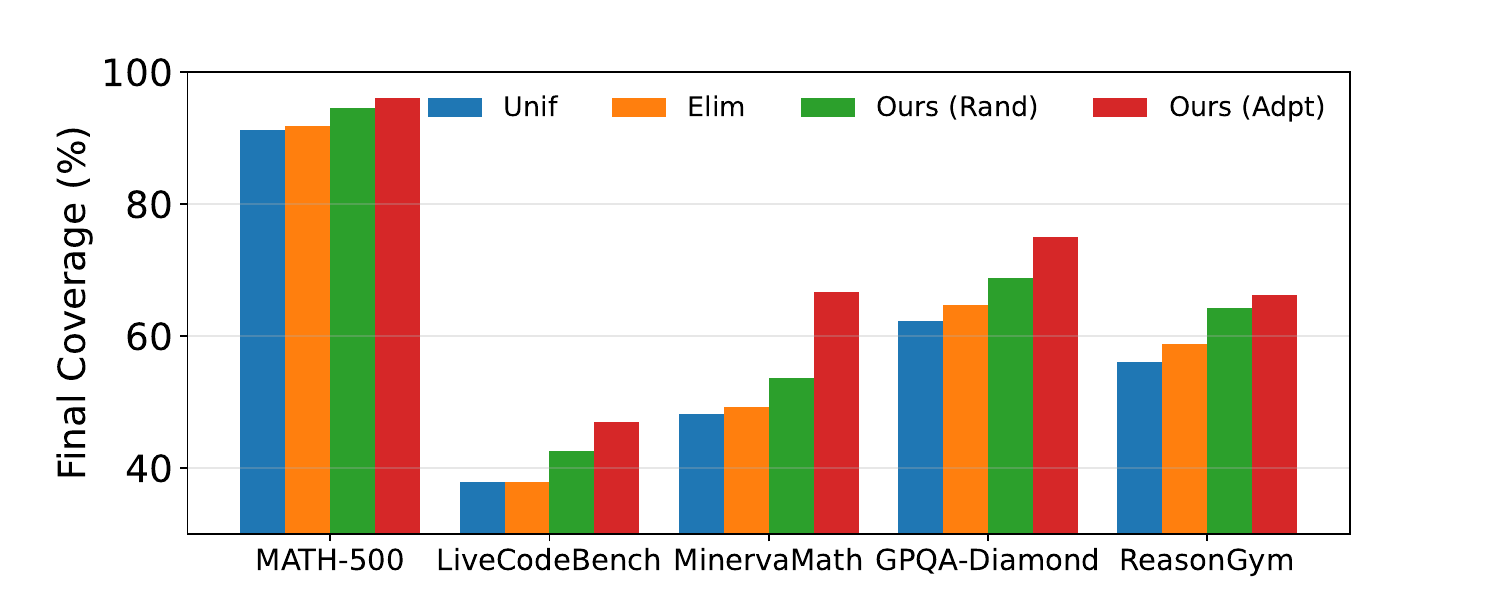}
    \caption{Gemini (non-thinking)}
  \end{subfigure}\hfill
  \begin{subfigure}[t]{0.48\linewidth}
    \centering
    \includegraphics[width=\linewidth]{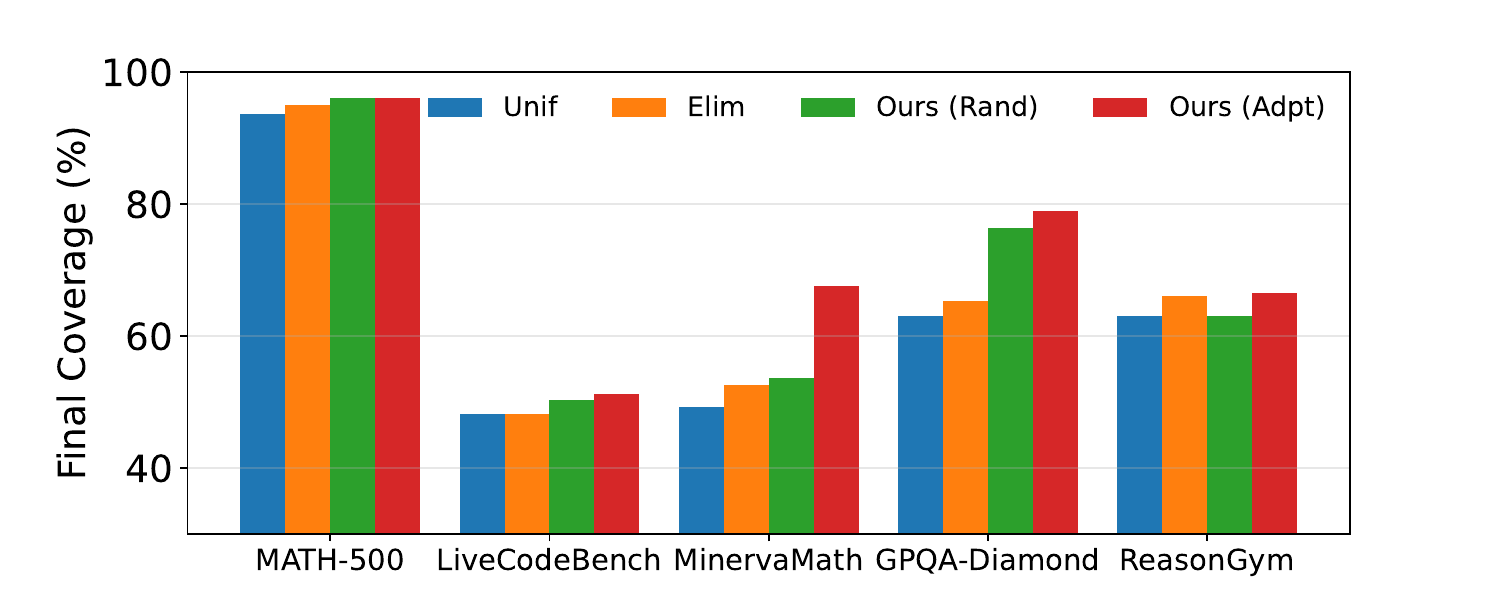}
    \caption{Gemini (thinking)}
  \end{subfigure}
   \begin{subfigure}[t]{0.48\linewidth}
    \centering
    \includegraphics[width=\linewidth]{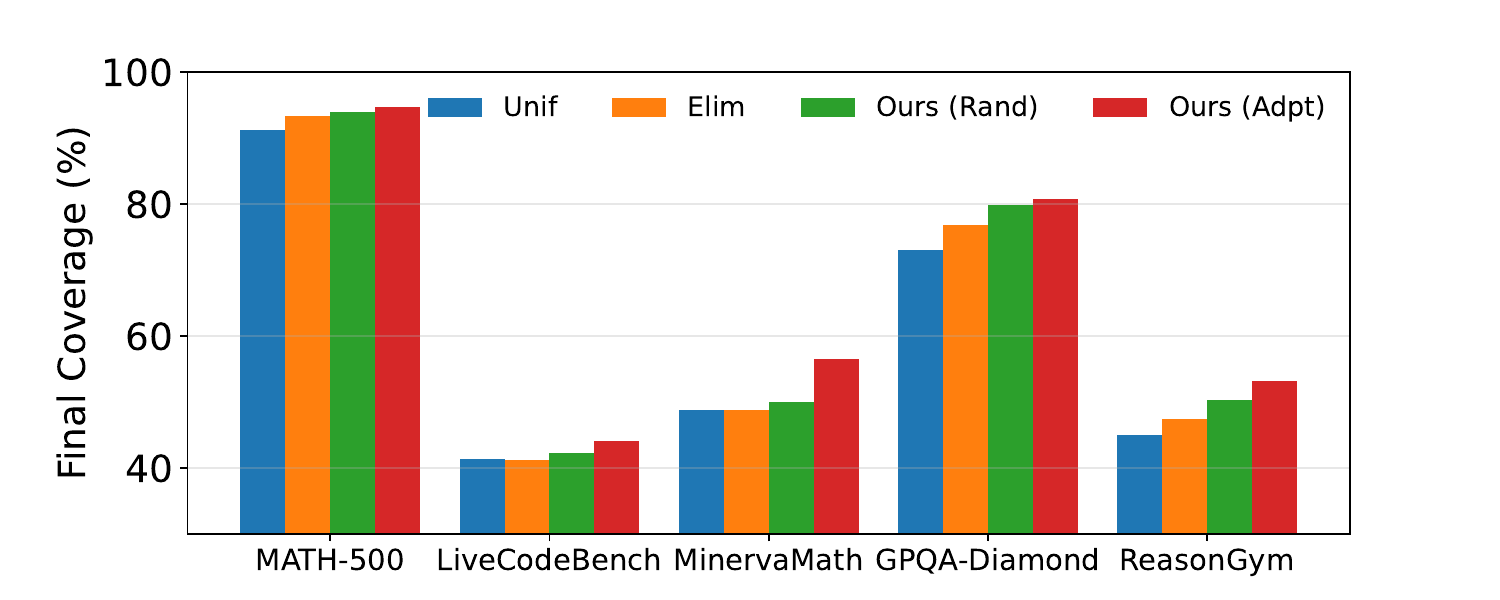}
    \caption{GPT-4 Mini}
  \end{subfigure}
  \begin{subfigure}[t]{0.48\linewidth}
    \centering
    \includegraphics[width=\linewidth]{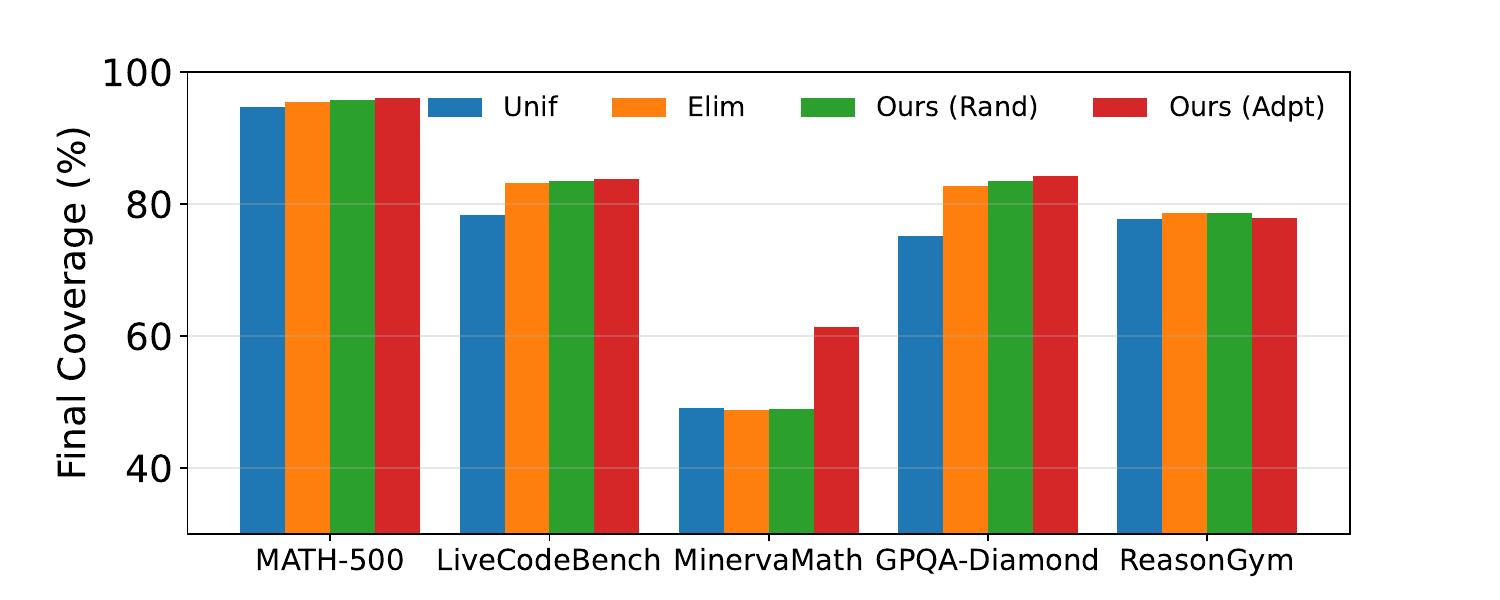}
    \caption{GPT-5 Nano}
  \end{subfigure}
  \caption{Final round coverage comparison}
  \label{fig:allhist}
\end{figure*}

\paragraph{Numerical results.}

We present numerical results for the approach \textsc{Ours (Adpt)} compared with all the other variants. We compare the last round coverage gain for \textsc{Ours (Adpt)} over the other baselines. In general, our method is able to achieve a good performance gain.

\begin{table}[t]
\caption{Coverage improvement (\%) for Gemini models. U/E/R = Uniform/Elimination/Random.}
\centering
\small
\setlength{\tabcolsep}{5pt}
\renewcommand{\arraystretch}{1.1}
\begin{tabular}{l ccc ccc}
\toprule
& \multicolumn{3}{c}{Nonthink} & \multicolumn{3}{c}{Think} \\
\cmidrule(lr){2-4} \cmidrule(lr){5-7}
Benchmark & U & E & R & U & E & R \\
\midrule
MATH-500      &  4.28 &  4.08 &  1.08 &  1.82 &  1.07 &  0.02 \\
LiveCodeBench &  7.43 &  7.43 &  3.46 &  2.56 &  2.56 &  0.20 \\
MinervaMath   & 15.93 & 15.36 & 11.86 & 16.11 & 14.43 & 13.39 \\
GPQA          & 11.07 & 10.23 &  7.03 & 12.16 & 11.40 &  1.68 \\
ReasonGym     &  8.98 &  8.06 &  1.52 &  2.27 &  0.58 &  3.38 \\
\bottomrule
\end{tabular}
\label{tab:gemini_coverage}
\end{table}

\begin{table}[t]
\caption{Coverage improvement (\%) for GPT models. U/E/R = Uniform/Elimination/Random.}
\centering
\small
\setlength{\tabcolsep}{5pt}
\renewcommand{\arraystretch}{1.1}
\begin{tabular}{l ccc ccc}
\toprule
& \multicolumn{3}{c}{GPT-4.1 Mini} & \multicolumn{3}{c}{GPT-5 Nano} \\
\cmidrule(lr){2-4} \cmidrule(lr){5-7}
Benchmark & U & E & R & U & E & R \\
\midrule
MATH-500      & 2.35 &  1.08 &  0.22 &  1.68 & 11.43 &  0.33 \\
LiveCodeBench & 2.55 &  3.11 &  2.01 &  3.41 &  1.14 &  1.67 \\
MinervaMath   & 7.14 &  7.41 &  6.34 & 10.29 & 11.43 & 11.34 \\
GPQA          & 5.50 &  3.25 &  0.67 &  5.56 &  1.68 &  1.05 \\
ReasonGym     & 6.31 &  5.04 &  2.33 &  0.31 & -0.02 &  0.31 \\
\bottomrule
\end{tabular}
\label{tab:gpt_coverage}
\end{table}

\section{Token efficiency analysis}
\label{app:token}
We attribute the token efficiency of our approach to two complementary factors:
(i) the characteristics of modern long-context LLMs, and
(ii) the ability to resolve queries that are effectively unsolvable under fixed distribution sampling.

\paragraph{Long context windows and token amplification.}
Modern LLMs—especially reasoning models \citep{deepseekai2025deepseekr1incentivizingreasoningcapability, comanici2025gemini25pushingfrontier}—operate with increasingly long context windows.
The explicit reasoning or ``thinking'' process alone can consume thousands of tokens, and generating a complete final response often requires additional substantial output.
As a result, a single failed attempt on a hard query can already incur a large token cost.
This issue is further exacerbated for difficult or ambiguous queries, where models may hallucinate extended reasoning chains without converging to a formatted answer \citep{xu2025hallucinationinevitableinnatelimitation, kalai2025languagemodelshallucinate}.
The same applies to powerful non-reasoning models, which often have large output budgets and can exhibit similar behavior when repeatedly sampling on challenging inputs.

Under a static sampling setting, hard queries tend to dominate the overall token budget: repeatedly allocating samples to such queries often leads to multiple long, unsuccessful generations, significantly inflating total token usage.
Consequently, naively increasing the number of samples does not necessarily improve efficiency and can instead amplify token consumption.

\paragraph{Solving unsolvable questions.}
The second—and more critical—factor behind token efficiency is the ability to actually resolve questions that are unlikely to be solved under static sampling.
Recent work by \citet{xia2025rethinkingunsolvableincontextsearch} shows that ICL can unlock solution paths for queries that appear unsolvable when sampled without ICL demonstrations.
By conditioning inference on carefully selected demonstrations, ICL can shift the model toward more effective reasoning trajectories, particularly for complex problems.
Our approach leverages this insight by incorporating ICL into our self-improving scaling scheme.
Once a hard question is successfully solved, further expensive generations for that question are avoided, leading to substantial savings in output tokens.
Importantly, the benefit is multiplicative: resolving a previously unsolvable question not only prevents repeated long hallucinated outputs for that question, but can also provide informative demonstrations that help guide inference on related queries.
Overall, token efficiency in our framework arises not merely from allocating fewer samples but from increasing the probability of early success on difficult queries.
By combining adaptive allocation with ICL-driven distributional shifts, our method reduces redundant long generations and makes more effective use of the large context windows available in modern LLMs.

\section{Failure mode}
To assess the robustness of similarity-based evolving ICL under weak semantic structure, we construct a deliberately heterogeneous test set. Specifically, we combine 60 questions drawn from LiveCodeBench, MATH500, and GPQA-Diamond (20 each), and further enforce diversity by selecting semantically distant instances based on cosine similarity in embedding space. This procedure intentionally disrupts local clustering and weakens nearest-neighbor relationships. The results are shown in \cref{tab:diverse_results}. While the absolute gains are smaller than those observed on more structured benchmarks, our method continues to yield consistent improvements over baseline strategies without any performance degradation. For example, at Round 3, \textsc{Ours (Adpt)} achieves 55\% coverage compared to 53.3\% for \textsc{Elim}. These results indicate that the benefits of evolving ICL are more limited in highly heterogeneous settings, though it still provides modest improvements over baseline approaches.

\begin{table}[t]
\caption{Coverage (\%) on a deliberately diverse dataset.}
\centering
\small
\setlength{\tabcolsep}{6pt}
\renewcommand{\arraystretch}{1.1}
\begin{tabular}{lcccc}
\toprule
Method & R0 & R1 & R2 & R3 \\
\midrule
Elim        & 45.0 & 53.3 & 53.3 & 53.3 \\
Ours (Rand) & 45.0 & 51.6 & 55.0 & 55.0 \\
Ours (Adpt) & 45.0 & 53.3 & 55.0 & 55.0 \\
\bottomrule
\end{tabular}
\label{tab:diverse_results}
\end{table}

\end{document}